\documentclass[letterpaper]{article} 
\usepackage[preprint]{aaai2027}  
\usepackage[hyphens]{url}  
\usepackage{graphicx} 
\usepackage{amsmath,amsfonts}
\usepackage{multirow}
\usepackage{natbib}  
\usepackage{caption} 
\usepackage{algorithm}
\usepackage{algorithmic}
\usepackage{booktabs}
\usepackage{enumitem}
\usepackage{siunitx}
\usepackage[most]{tcolorbox}
\tcbuselibrary{breakable,skins}

\newtcolorbox{promptbox}[1][]{
    enhanced jigsaw,
    breakable,
    title={English Prompt},
    colback=white,
    colframe=black!70,
    boxrule=0.5pt,
    arc=1mm,
    left=1.5mm,
    right=1.5mm,
    top=1mm,
    bottom=1mm,
    before skip=6pt,
    after skip=6pt,
    fonttitle=\bfseries,
    fontupper=\small,
    #1
}

\usepackage{newfloat}
\usepackage{listings}
\DeclareCaptionStyle{ruled}{labelfont=normalfont,labelsep=colon,strut=off} 
\floatstyle{ruled}
\newfloat{listing}{tb}{lst}{}
\floatname{listing}{Listing}

\usepackage{booktabs}

\title{\textsc{FirmCORe}: A Benchmark for Structured Reasoning about Inter-Firm Collaboration Opportunities}
\author{
    Tian Du\textsuperscript{\rm 1,2,}\corresponding,
    Tiantong Wu\textsuperscript{\rm 2},
    Yafei Wang\textsuperscript{\rm 3},
    Mengyu Liu\textsuperscript{\rm 1},
    Xingyan Chen\textsuperscript{\rm 3},
    Mu Wang\textsuperscript{\rm 3}
}
\affiliations{
    \textsuperscript{\rm 1}Southwest University of Finance and Economics\\
    \textsuperscript{\rm 2}Nanyang Technological University\\
    \textsuperscript{\rm 3}Beijing University of Post and Telecommunication
}

\begin{document}

\maketitle

\begin{abstract}

Comprehensive structured data on inter-firm relationships is often scarce or inaccessible because many relationships are privately negotiated, selectively disclosed, and fragmented across proprietary databases. This scarcity hinders the discovery of collaboration opportunities, particularly for startups and small and medium-sized enterprises. Firm profiles are readily available, but collaboration potential cannot be inferred from business similarity alone, since similar firms may be competitors, whereas dissimilar firms may offer complementary products, technologies, channels, capabilities, or capital. 
We present \textsc{FirmCORe} (Inter-\underline{\textbf{Firm}} \underline{\textbf{C}}ollaboration \underline{\textbf{O}}pportunity \underline{\textbf{Re}}asoning), a human-annotated benchmark for pairwise reasoning over weakly structured firm profiles, comprising 2,805 labeled firm pairs. Given two firm profiles, a model must determine whether the available evidence supports a collaboration opportunity and, for positive pairs, jointly predict its strength, primary collaboration type, and role direction. \textsc{FirmCORe} also provides parallel Chinese- and English-language evaluation sets containing identical instances and gold labels, enabling controlled analysis of input-language sensitivity. Experiments with representative locally deployed and hosted large language models (LLMs) show that the strongest model achieves a macro-F1 score of 74.51 for opportunity detection but only 61.57\% exact match across all four output fields. Language effects vary across models, and high cross-language agreement can mask errors shared across languages. These results indicate that current LLMs are substantially more reliable at detecting broad collaboration opportunities than at identifying their specific types and role directions.

\end{abstract}


\section{Introduction}
\label{sec:introduction}
Identifying potential inter-firm collaboration opportunities is important for industrial coordination and supply-chain resilience. However, comprehensive structured data on inter-firm relationships remains scarce: many collaborations are established through private negotiations, disclosed only selectively, or recorded in proprietary databases. This data gap is particularly severe for startups and small and medium-sized enterprises, which often lack the resources to identify suitable suppliers, customers, and other potential collaborators. Existing transaction networks and structured supply–demand databases capture only a subset of realized relationships and provide even less information about collaboration opportunities that have not yet materialized. In contrast, firm profiles, including core business activities, products and services, and profile descriptions, are more widely available and therefore provide a practical source of evidence.

Inferring collaboration opportunities from firm profiles is nevertheless challenging. Such profiles are often incomplete, weakly structured, and expressed at varying levels of granularity using heterogeneous or domain-specific terminology. More fundamentally, collaboration depends on capability complementarity rather than firm similarity~\cite{hitt2000partner,furlotti2018fit}. Similar firms may be competitors, whereas dissimilar firms may complement one another through products, technologies, distribution channels, operational capabilities, or capital~\cite{mitsuhashi2009matching,mindruta2016two,greve2013greener}. Reliable reasoning must therefore determine not only whether the available profile evidence supports a collaboration opportunity, but also its opportunity strength, primary collaboration type, and the role direction of the two firms.

Existing research does not directly evaluate this form of reasoning. Corporate similarity methods primarily measure semantic or business relatedness~\cite{davis2022machine,almahri2026enhancing}. Supply-chain link prediction typically relies on observed relational graphs~\cite{cabrera2021approach,tu2024using}. Partner recommendation is usually designed for candidate retrieval or ranking in specific application settings. None jointly evaluates whether two firm profiles provide sufficient evidence for a potential collaboration.
Large language models (LLMs) offer a promising approach because they can integrate heterogeneous textual evidence and interpret domain-specific terminology~\cite{almahri2026enhancing,zheng2025enhancing,sun2025intercorprel}. However, they may conflate relatedness with complementarity, infer plausible but weakly supported collaboration opportunities, or produce inconsistent predictions across opportunity existence, opportunity strength, collaboration type, and role direction. Consequently, the ability of LLMs to perform multidimensional inter-firm collaboration reasoning from firm profiles remains largely unknown.

\begin{figure*}[!ht]
  \centering
  \includegraphics[width=\linewidth]{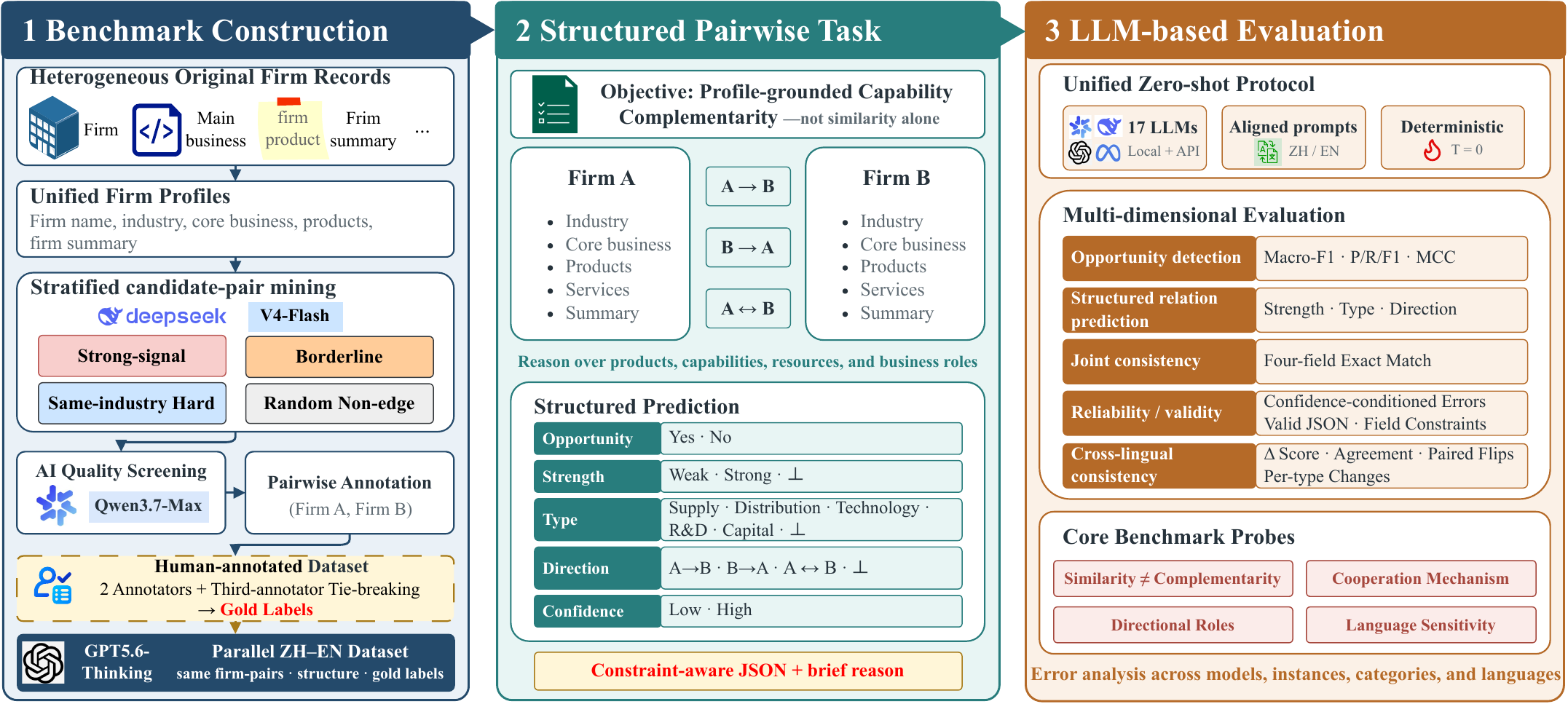}
  \caption{Overview of \textsc{FirmCORe}. The benchmark is constructed from unified firm profiles through stratified candidate mining, quality screening, blind human annotation, and tuple-level adjudication. Models then perform zero-shot structured prediction on parallel Chinese and English profile pairs under a unified evaluation protocol.}
  \label{fig:framework}
\end{figure*}

We investigate three research questions: (\textbf{RQ1}) Can LLMs distinguish evidence-supported collaboration opportunities from superficial business relatedness? (\textbf{RQ2}) Can they jointly recover opportunity strength, collaboration type, and role direction? (\textbf{RQ3}) Would the structured predictions remain stable across firm profiles in different languages? Additional analyses of model output validity, consistency across different prediction fields, and the model’s self-reported confidence were also included.

To address these RQs, we introduce \textsc{FirmCORe} (Inter-\underline{\textbf{Firm}} \underline{\textbf{C}}ollaboration \underline{\textbf{O}}pportunity \underline{\textbf{Re}}asoning), a human-annotated benchmark comprising 2,805 gold-labeled firm pairs for structured reasoning about inter-firm collaboration opportunities. 
The main contributions of this paper are as follows:
\begin{itemize}[leftmargin=*]

\item We formulate inter-firm collaboration opportunity detection as a profile-based, capability-complementarity reasoning task that jointly predicts opportunity existence, opportunity strength, collaboration type, and role direction.

\item We introduce \textsc{FirmCORe}, a human-annotated benchmark containing 2,805 gold-labeled firm pairs derived from real-world firm profiles, together with parallel Chinese and English evaluation sets.

\item We systematically evaluate representative LLMs under a unified structured prediction setting and show that detecting broad collaboration opportunities is substantially easier than correctly identifying collaboration types, role directions, and complete four-field relation structures.

\end{itemize}

\section{Related Work}

\textbf{Firm Similarity and Representation.}
Prior work uses business descriptions, industry attributes, and inter-firm graphs to support tasks such as similarity estimation, competitor retrieval, and industry classification \cite{cao2024companykg,yang2025emerging}. These methods primarily capture semantic or structural relatedness between firms. However, similarity in products, business activities, or industry affiliation does not necessarily indicate complementary capabilities or an evidence-supported collaboration opportunity.

\textbf{Link Prediction and Supplier Recommendation.}
Supply-chain link prediction and supplier recommendation leverage relational graphs, knowledge bases, firm attributes, and procurement signals to predict supplier--customer links or rank candidate firms for collaboration~\cite{kosasih2025towards,li2025integrating}. The former relies primarily on observed inter-firm network structure, whereas the latter typically operates within specific platforms or application contexts. In contrast, \textsc{FirmCORe} uses only two weakly structured firm profiles, without relying on inter-firm networks or platform interactions, to distinguish capability complementarity from superficial relatedness.

\textbf{LLM-Based Benchmarks for Business Reasoning.}
Recent benchmarks evaluate LLMs on economic reasoning, corporate tasks, and supply-chain knowledge or workflow execution~\cite{quan2024econlogicqa,guan2026supchain}, primarily through individual questions, documents, simulated scenarios, or predefined operational workflows~\cite{krumdick2024bizbench,matlin2025finance}. \textsc{FirmCORe} instead targets profile-grounded structured reasoning about inter-firm collaboration opportunities. Given two weakly structured firm profiles, it assesses whether LLMs can detect an evidence-supported collaboration opportunity and, for positive pairs, predict opportunity strength, collaboration type, and role direction.

\section{The \textsc{FirmCORe} Benchmark}
\label{sec:benchmark}

\subsection{Task Definition}

Each benchmark instance consists of a pair of firm profiles presented in a fixed randomized order as Firm A and Firm B. Each firm profile includes the firm name, industry, core business, products and services, and a brief profile summary. Each firm pair appears only once, and the reversed ordering is not included separately. For each pair $(i,j)$, \textsc{FirmCORe} provides the following structured label:
\begin{equation}
    y_{ij}=(o_{ij}, g_{ij}, t_{ij}, r_{ij}),
\end{equation}
where $o_{ij}$ indicates whether a collaboration opportunity exists, $g_{ij}$ denotes its strength, $t_{ij}$ denotes the primary collaboration type, and $r_{ij}$ denotes the role direction.

Opportunity existence is binary: \(o_{ij} \in \{0,1\}\), where \(o_{ij}=1\) indicates that a collaboration opportunity exists, and \(o_{ij}=0\) otherwise. Conditional on \(o_{ij}=1\), the remaining labels satisfy: \(g_{ij} \in \{\textsc{Weak}, \textsc{Strong}\}\), \(t_{ij} \in \{\textsc{Supply}, \textsc{Distribution},
\textsc{Technology}, \textsc{R\&D}, \textsc{Capital}\}\), and 
\(r_{ij} \in \{\textsc{A2B}: A\leftarrow B, \textsc{B2A}: B\leftarrow A,
\textsc{Bidirectional}: A\leftrightarrow B\}\).
\(g_{ij}=\textsc{Strong}\) denotes a direct and well-supported collaboration interface, whereas \(g_{ij}=\textsc{Weak}\) denotes a plausible but indirect or less clearly supported opportunity. The collaboration type represents the most direct and best-supported form of collaboration. Role direction is defined relative to the displayed A/B order and captures the primary flow of resources or capabilities.

\textsc{FirmCORe} evaluates whether the supplied profiles support a potential collaboration, rather than whether the firms have collaborated in practice. Models must rely solely on the provided profiles. They also generate a brief rationale and a \textsc{Low}/\textsc{High} confidence label, which are analyzed separately from four-field exact match.

\subsection{\textsc{FirmCORe} Framework}

Figure~\ref{fig:framework} presents an overview of the \textsc{FirmCORe} framework, which is structured into three stages and five key components.
\paragraph{A. Firm Profile Construction} 
We consolidate records from heterogeneous data sources into unified firm profiles, each containing the firm name, industry, core business, products and services, and a textual summary. Records are matched based on identifiers and names, duplicate values are merged at the field level, and missing fields are omitted without imputation. We then remove records lacking valid firm names, deduplicate the resulting firm profiles, and exclude profiles that are too sparse for meaningful pairwise assessment. Data provenance, temporal and geographic coverage, privacy filtering, and release conditions are documented in Appendix~\ref{app:data-statement}.

\paragraph{B. Stratified Candidate Pair Mining} To prevent the benchmark from being dominated by trivial negative samples, we mine candidate firm pairs from four strata: strong-signal pairs, borderline-signal pairs, same-industry hard pairs that exhibit business relatedness but lack explicit supply--demand complementarity, and randomly sampled non-edge pairs. Upstream relationship scores are computed from firm profiles and true available supply--demand information and are used solely for candidate selection, never as ground-truth labels. We retain up to 3{,}000 firm pairs per stratum and remove duplicate pairs as well as pairs that differ only in ordering. We use DeepSeek-V4-Flash to screen for potential label leakage, privacy risks, and data-quality issues. Up to 750 pairs per stratum are selected for human annotation, while stratum membership, associated scores, and source identifiers are hidden from annotators. Detailed mining criteria and thresholds are provided in Appendix~\ref{app:candidate_mining}.

\paragraph{C. Human Annotation}
Each firm pair was independently annotated by two annotators under source-blind conditions~\cite{bender2018data}. Annotators relied solely on the provided firm profiles and were not allowed to consult external sources. They labeled opportunity existence and, for positive pairs, opportunity strength, collaboration type, and role direction. A positive label required an evidence-supported collaboration interface involving products, services, technologies, channels, capabilities, resources, or capital. Industry similarity or lexical overlap alone was insufficient. For negative pairs, the remaining three fields were set to \textsc{None} ($\perp$). 

\paragraph{D. Gold Label Aggregation} Labels were aggregated at the level of complete four-field tuples rather than by field-wise voting, which could produce internally inconsistent combinations that no annotator had selected. If the two initial annotators disagreed on the complete four-field tuple, a third annotator independently labeled the instance as a tie breaker. The gold tuple was accepted only when the third annotation matched one of the two initial
tuples; cases with three distinct tuples or insufficient evidence remained unresolved and were excluded. Third-annotator adjudication was required for 30.55\% of the samples. This process yielded 2,805 gold-standard instances, while the remaining 195 cases were excluded because the disagreements could not be resolved with sufficient confidence. The two initial annotators achieved an average field-level agreement of 84.26\% and a full-tuple agreement of 69.45\%. See Table~\ref{tab:statistics} for detailed statistics.

\begin{table}[t]
\centering
\small
\caption{Statistics of the \textsc{FirmCORe} benchmark.}
\label{tab:statistics}
\begin{tabular}{@{}lr@{}}
\toprule
Statistic & Value \\
\midrule
\multicolumn{2}{@{}l}{\textit{Firm profiles}} \\
Firms / industries & 3{,}230 / 1{,}114 \\
Avg. profile length (characters) & 179.09 \\
\midrule
\multicolumn{2}{@{}l}{\textit{Annotation outcomes}} \\
Gold / unresolved pairs & 2{,}805 / 195 \\
Positive / negative pairs & 968 / 1{,}837 \\
Weak / strong opportunities & 35 / 933 \\
\midrule
\multicolumn{2}{@{}l}{\textit{Collaboration types}} \\
Supply / distribution & 641 / 114 \\
Technology / R\&D & 86 / 112 \\
Capital & 15 \\
\midrule
\multicolumn{2}{@{}l}{\textit{Role directions}} \\
A2B / B2A / bidirectional & 157 / 700 / 111 \\
\midrule
\multicolumn{2}{@{}l}{\textit{Annotation quality}} \\
Field / tuple agreement & 84.26\% / 69.45\% \\
Third-annotator involvement & 30.55\% \\
\midrule
\multicolumn{2}{@{}l}{\textit{Parallel ZH--EN sets}} \\
Chinese / English instances & 2{,}805 / 2{,}805 \\
Pairs flagged for review & 144 \\
\bottomrule
\end{tabular}
\begin{minipage}{0.95\columnwidth}
\footnotesize
\textit{Note.} Distribution denotes Marketing and Distribution;
Technology denotes Technology Transfer; R\&D denotes R\&D and
Co-development.
\end{minipage}
\end{table}

\paragraph{E. Structured Model Evaluation}
\label{sec:evaluation}
Let $\ell \in \{\mathrm{zh},\mathrm{en}\}$ denote the input language. 
Each evaluation instance is expressed as
\begin{equation}
d_{ij}^{\ell}
=
\left(
x_i^{\ell},
x_j^{\ell},
y_{ij}^{*}
\right),
y_{ij}^{*}
=
\left(
o_{ij}^{*},
g_{ij}^{*},
t_{ij}^{*},
r_{ij}^{*}
\right),
\end{equation}
where $x_i^{\ell}$ and $x_j^{\ell}$ are the firm profiles presented as
Firm A and B, respectively. The gold structured label
$y_{ij}^{*}$ contains opportunity existence, opportunity strength,
collaboration type, and role direction. Given a language-aligned task instruction $p^{\ell}$, a model
$f_{\theta}$ produces a raw textual response
\begin{equation}
z_{ij}^{\ell}
=
f_{\theta}
\left(
p^{\ell},
x_i^{\ell},
x_j^{\ell}
\right).
\end{equation}
A deterministic parser $\Pi$ maps the raw response to
\begin{equation}
\Pi\left(z_{ij}^{\ell}\right)
=
\left(
\hat{y}_{ij}^{\ell},
\hat{c}_{ij}^{\ell},
\hat{e}_{ij}^{\ell}
\right), 
\hat{y}_{ij}^{\ell}
=
\left(
\hat{o}_{ij}^{\ell},
\hat{g}_{ij}^{\ell},
\hat{t}_{ij}^{\ell},
\hat{r}_{ij}^{\ell}
\right)
\end{equation}
where $\hat{y}_{ij}^{\ell}$ is the predicted four-field label, $\hat{c}_{ij}^{\ell}$ is the
self-reported confidence, and $\hat{e}_{ij}^{\ell}$ is a brief rationale. A structurally valid prediction must satisfy
\begin{equation}
\hat{o}_{ij}^{\ell}=\textsc{No}
\quad\Longrightarrow\quad
\hat{g}_{ij}^{\ell}
=
\hat{t}_{ij}^{\ell}
=
\hat{r}_{ij}^{\ell}
=
\textsc{None}.
\end{equation}

Let $\mathcal{I}^{\ell}$ denote the set of all evaluation instances and let $\mathcal{I}_{+}^{\ell}
=
\left\{
(i,j)\in\mathcal{I}^{\ell}
\,\middle|\,
o_{ij}^{*}=\textsc{Yes}
\right\}$
denote the gold-positive subset. Opportunity detection is evaluated on all instances. Opportunity strength, collaboration type, and role direction are
evaluated on $\mathcal{I}_{+}^{\ell}$.

Opportunity strength, collaboration type, and role direction
are evaluated end-to-end on all gold-positive instances
$\mathcal{I}^{\ell}_{+}$. If a model predicts \textsc{No}, or
produces an invalid or missing relation attribute, the
corresponding attribute prediction is counted as incorrect.
Oracle-detection evaluation instead conditions on the subset
of gold-positive instances that the model correctly detects
as opportunities. The former measures complete pipeline
performance, whereas the latter isolates relation-attribute
reasoning after successful detection.
For full structured prediction, we use four-field exact match:
\begin{equation}
\mathrm{EM}^{\ell}
=
\frac{1}{|\mathcal{I}^{\ell}|}
\sum_{(i,j)\in\mathcal{I}^{\ell}}
\mathbb{I}
\left[
\hat{y}_{ij}^{\ell}=y_{ij}^{*}
\right],
\end{equation}
where $\mathbb{I}[\cdot]$ denotes the indicator function. For the parallel Chinese and English evaluation sets, cross-lingual consistency is measured by paired exact agreement:
\begin{equation}
\mathrm{Agree}_{\mathrm{zh,en}}
=
\frac{1}{|\mathcal{I}|}
\sum_{(i,j)\in\mathcal{I}}
\mathbb{I}
\left[
\hat{y}_{ij}^{\mathrm{zh}}
=
\hat{y}_{ij}^{\mathrm{en}}
\right].
\end{equation}

\subsection{Chinese--English Parallel Evaluation Set}

To assess model sensitivity to input language under controlled conditions, we constructed an English parallel evaluation set~\cite{han2026mubench}. It contains the same 2,805 firm pairs, profile fields, and gold labels as the original Chinese benchmark. Let $\mathcal{P}$ denote the fixed set of annotated ordered firm pairs,
and let $x_k^{\mathrm{zh}}$ denote the Chinese profile of firm $k$.
For each language $\ell\in\{\mathrm{zh},\mathrm{en}\}$, we define
\begin{equation}
\mathcal{D}^{\ell}
=
\left\{
d_{ij}^{\ell}
\,\middle|\,
(i,j)\in\mathcal{P}
\right\},
x_k^{\mathrm{en}}=T\!\left(x_k^{\mathrm{zh}}\right),
\label{eq:parallel-en}
\end{equation}
where $T(\cdot)$ denotes the field-level translation function.  $\mathcal{D}^{\mathrm{zh}}$ and $\mathcal{D}^{\mathrm{en}}$
contain identical ordered firm pairs and gold labels, differing only
in the language of the firm profiles. More details are shown in Appendix~\ref{app:zh-en}.

The English profiles were translated field by field with assistance from GPT5.6-Thinking and then selectively reviewed by human annotators. The translation process was designed to preserve the semantic scope, level of detail, and uncertainty of the Chinese originals. For recurring companies, consistent translations were used for firm names, industries, core business descriptions, products and services, and profile summaries. We flagged 94 distinct firm profiles appearing in 144 firm-pair instances for mandatory review because they involved potential ambiguities in terminology, firm names, abbreviations, business roles, or directional semantics. Annotators compared each flagged English field against its Chinese source and corrected any errors. This paired design enables instance-level comparison of model predictions while holding sample composition and gold labels constant. The complete Chinese and English prompts, output schema, and label mappings are provided in Appendix~\ref{app:prompt}.

\section{Experiments}

\subsection{Experimental Setups} 
\paragraph{Models} We evaluate 17 LLMs, ranging from 1.5B-parameter models to large mixture-of-experts systems. The benchmark covers both locally deployed open-weight models, including DeepSeek-7B, Llama3.1-8B, Llama3.2-3B, Qwen3-2B/4B/8B/Coder, Gemma3-12B, and Gemma4-26B, and hosted models, including GLM-5.1/5.2, Kimi-K2.7, Qwen3-VL-8B, Qwen3.6-27B, Qwen3.6-35B, Qwen3.6-Max, and Qwen3.7-Plus.

\paragraph{Implementation}

 Each model received profiles of Firm A and B and a standardized prompt, and was asked to determine whether a collaboration opportunity existed, its strength and type, the directional roles, and the model’s self-reported confidence. All models used the same prompt template, output parser, label-normalization pipeline, and evaluation scripts. All models were evaluated in a zero-shot setting, with no examples or task-specific fine-tuning. Provider-specific reasoning modes were disabled. Each sample was evaluated once with a temperature of 0 and a maximum output length of 2,048 tokens. Hosted models were accessed through OpenAI-compatible chat completion APIs with a 180-second timeout. Transient connection failures, rate-limit errors, and server errors were retried up to four times. Local models were evaluated using Hugging Face Transformers. 

\paragraph{Evaluation Metrics}
We follow the structured evaluation protocol defined in Section~\ref{sec:evaluation}. Opportunity detection is primarily measured by Macro-F1. Relation attributes are evaluated under both end-to-end and oracle-detection settings and complete predictions are assessed using four-field exact match. We also report paired Chinese–English agreement, output validity, and confidence-based reliability metrics.

\section{Results and Analysis}

Following the RQs, we first evaluate opportunity detection, then examine structured relation recovery and input-language sensitivity. We conclude with an analysis of output validity and self-reported confidence. Additional baseline comparisons, complete diagnostic results, and challenging failure cases are
provided in Appendix~\ref{app:additional-results}.

\subsection{RQ1: Distinguishing Collaboration Opportunities from Relatedness}

Table~\ref{tab:opportunity-detection} reports performance on the collaboration opportunity detection. \textbf{LLMs tend to conflate business relatedness with genuine collaboration potential.} Qwen3.6-27B performs best, achieving a Macro-F1 of 74.51 with a 95\% bootstrap confidence interval of 72.89–76.07. Its positive-class precision and recall are 59.99\% and 84.71\%, respectively. Several other models attain near-saturated recall but substantially lower precision. For example, Qwen3.6-35B and GLM-5.1 reach recalls of 97.62\% and 98.86\%, yet their precision is only 44.16\% and 38.90\%. This pattern indicates systematic overprediction once a model detects a seemingly plausible industrial or commercial connection.

\begin{table}[hb!]
\centering
\caption{Opportunity detection performance on Chinese.}
\label{tab:opportunity-detection}
\small
\setlength{\tabcolsep}{3.8pt}
\renewcommand{\arraystretch}{0.98}
\begin{tabular}{lccccc}
\toprule
\multirow{2}{*}{Model}
& \multirow{2}{*}{Macro-F1}
& \multicolumn{3}{c}{Positive}
& \multirow{2}{*}{MCC} \\
\cmidrule(lr){3-5}
& & P & R & F1 & \\
\midrule
Qwen3-2B      & 51.50 & 38.10 & 26.96 & 31.58 & 4.30 \\
Qwen3-8B      & 51.42 & 42.49 & 93.80 & 58.49 & 28.55 \\
Gemma4-26B    & 50.82 & 41.69 & 97.93 & 58.48 & 31.25 \\
DeepSeek-7B   & 37.64 & 36.69 & 94.73 & 52.90 & 12.96 \\
Llama3.1-8B   & 36.50 & 36.87 & 98.97 & 53.73 & 17.62 \\
Qwen3-4B      & 34.48 & 36.34 & 97.93 & 53.01 & 14.20 \\
Gemma3-12B    & 32.22 & 35.93 & 99.90 & 52.86 & 14.56 \\
Llama3.2-3B   & 25.77 & 34.65 & 99.69 & 51.43 & 1.89 \\
\midrule
Qwen3.6-27B   & \textbf{74.51} & 59.99 & 84.71 & 70.24 & 52.25 \\
Qwen3.6-Max   & \underline{70.23} & 54.40 & 89.46 & 67.66 & 47.88 \\
Qwen3.7-Plus  & 69.64 & 53.62 & 91.84 & 67.71 & 48.33 \\
Qwen3-Coder   & 56.48 & 43.75 & 89.98 & 58.87 & 30.39 \\
Qwen3.6-35B   & 56.04 & 44.16 & 97.62 & 60.81 & 36.37 \\
GLM-5.2       & 53.78 & 42.97 & 97.21 & 59.59 & 33.57 \\
Kimi-K2.7     & 52.87 & 42.55 & 97.42 & 59.23 & 32.87 \\
Qwen3-VL-8B   & 44.46 & 39.32 & 96.80 & 55.92 & 23.49 \\
GLM-5.1       & 43.22 & 38.90 & 98.86 & 55.83 & 24.67 \\
\bottomrule
\end{tabular}

\begin{minipage}{0.98\columnwidth}
\footnotesize
Scores are percentages. $N=2{,}805$ (968 positive, 1,837 negative).
Incomplete prediction files are excluded, 95\% CIs are provided in the machine-readable results.
\end{minipage}
\end{table}

\begin{figure}[ht!]
    \centering
    \includegraphics[width=\columnwidth]
    {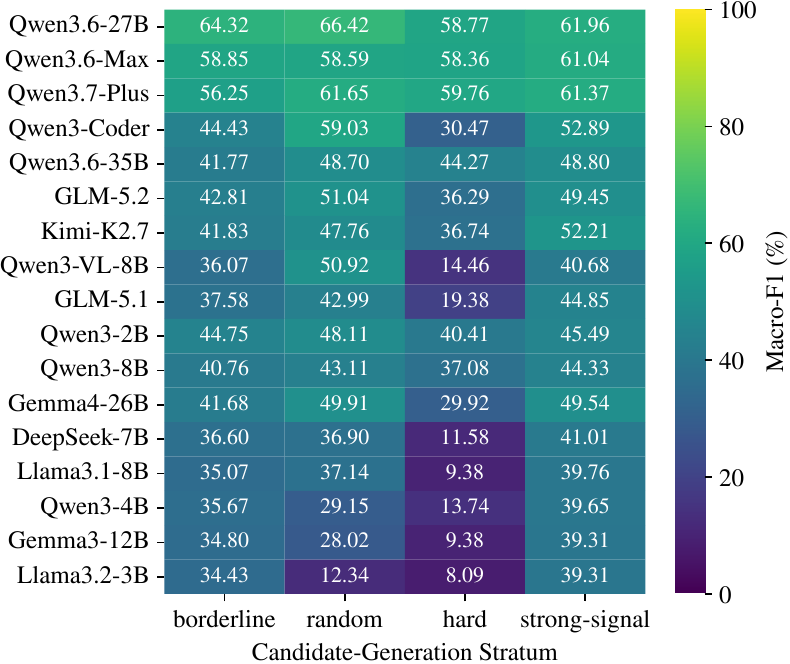}
    \caption{Opportunity-detection Macro-F1 by candidate-generation stratum on the Chinese benchmark. Stratum prevalence differs substantially, so cross-stratum values are descriptive rather than controlled estimates of intrinsic difficulty.}
    \label{fig:rq1-stratum}
\end{figure}

\begin{table}[ht!]
\centering
\small
\setlength{\tabcolsep}{3.8pt}
\renewcommand{\arraystretch}{0.98}
\caption{End-to-end structured relation prediction on the Chinese benchmark. F1 denotes Macro-F1.}
\label{tab:structured-relation}
\begin{tabular}{lcccccc}
\toprule
\multirow{2}{*}{Model}
& \multicolumn{2}{c}{Strength}
& \multicolumn{2}{c}{Type}
& \multicolumn{2}{c}{Direction} \\
\cmidrule(lr){2-3}
\cmidrule(lr){4-5}
\cmidrule(lr){6-7}
& {Acc.} & {F1}
& {Acc.} & {F1}
& {Acc.} & {F1} \\
\midrule
Qwen3-2B
& 20.66 & 19.59
& 15.08 & 13.05
& 7.64 & 13.34 \\

Qwen3-8B
& 69.94 & 46.42
& 67.56 & 47.42
& 69.11 & 52.10 \\

Gemma4-26B
& 66.53 & 46.00
& 70.66 & 53.34
& 73.45 & 66.52 \\

DeepSeek-7B
& 78.51 & 46.41
& 32.54 & 31.75
& 38.95 & 36.31 \\

Llama3.1-8B
& 77.48 & 45.93
& 60.02 & 42.77
& 67.56 & 52.25 \\

Qwen3-4B
& 86.88 & 50.73
& 68.70 & 57.14
& 59.71 & 54.89 \\

Gemma3-12B
& 68.08 & 45.27
& 51.65 & 44.44
& 63.84 & 61.02 \\

Llama3.2-3B
& 55.48 & 38.85
& 46.69 & 14.55
& 31.71 & 22.25 \\

\midrule
Qwen3.6-27B
& 64.67 & 44.62
& 60.12 & 51.05
& 63.95 & 56.11 \\

Qwen3.6-Max
& 76.34 & 49.87
& 66.32 & 51.97
& 68.39 & 58.61 \\

Qwen3.7-Plus
& 72.42 & 46.74
& 68.18 & \underline{58.10}
& 72.73 & 67.04 \\

Qwen3-Coder
& 86.98 & 46.70
& 64.67 & 46.65
& 68.08 & 59.54 \\

Qwen3.6-35B
& 75.62 & 48.04
& 70.87 & 54.53
& 68.90 & 63.38 \\

GLM-5.2
& 75.21 & 50.00
& 67.56 & 54.25
& 67.87 & 59.87 \\

Kimi-K2.7
& 88.64 & 53.08
& 74.28 & \textbf{60.23}
& 77.38 & 67.71 \\

Qwen3-VL-8B
& 60.85 & 41.83
& 59.61 & 40.07
& 68.08 & 54.61 \\

GLM-5.1
& 58.16 & 42.32
& 65.50 & 56.28
& 71.80 & 63.72 \\
\bottomrule
\end{tabular}
\end{table}

\begin{figure}[ht!]
  \centering
  \includegraphics[width=\columnwidth]{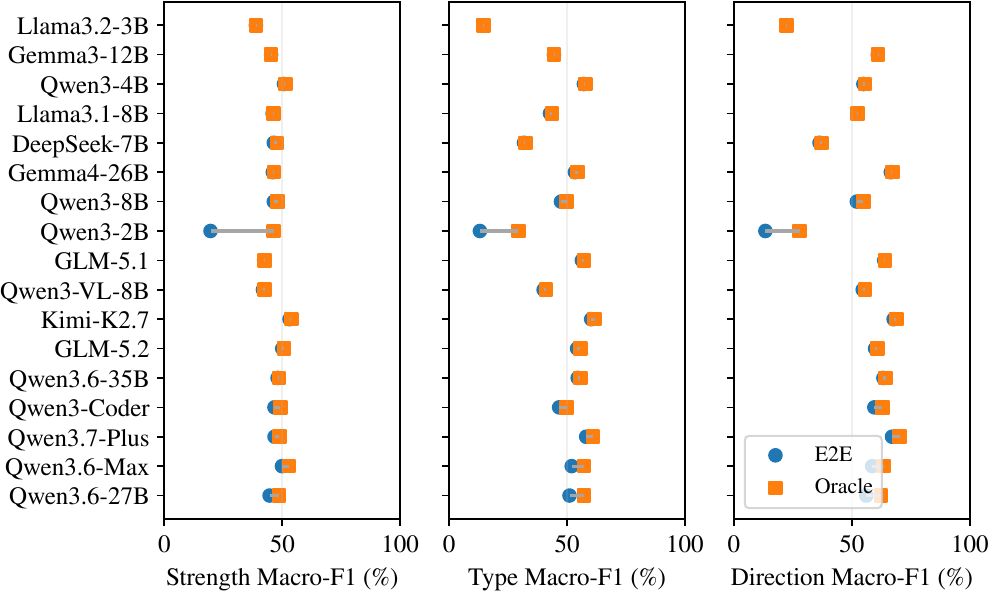}
  \caption{End-to-end and oracle-detection Macro-F1 for opportunity
  strength, collaboration type, and role direction on the Chinese benchmark.
  End-to-end evaluation retains all gold-positive instances and counts a
  missed opportunity as an attribute error; oracle evaluation conditions on
  correctly detected gold-positive instances.}
  \label{fig:rq2-e2e-oracle}
\end{figure}

\begin{figure}[ht!]
    \centering
    \includegraphics[width=\columnwidth]
    {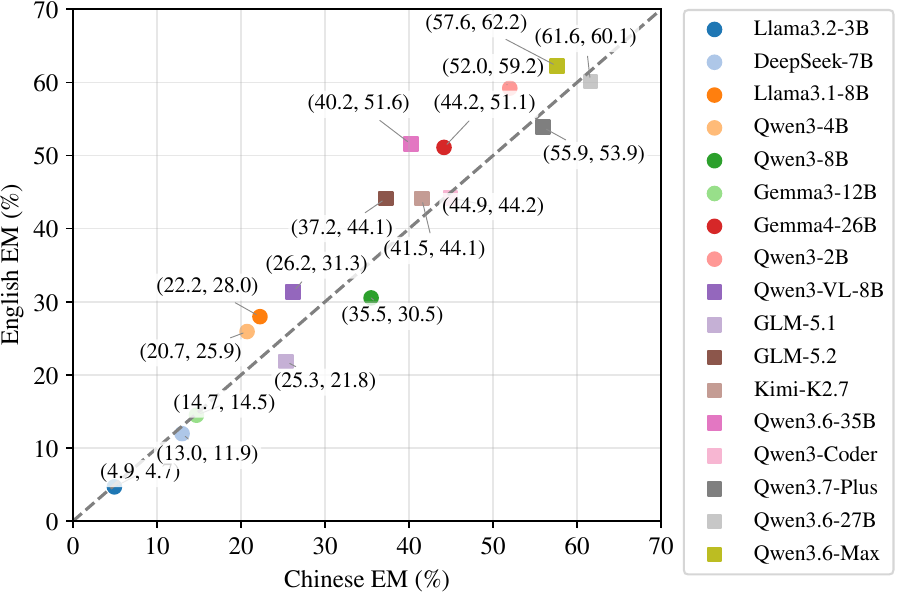}
    \caption{Chinese versus English four-field exact match across evaluated models. The dashed diagonal indicates equal performance in both languages.}
    \label{fig:rq3-zh-en}
\end{figure}

Figure~\ref{fig:rq1-stratum} further shows that models generally perform worse on same-industry hard pairs. For Qwen3.6-27B, Macro-F1 drops from 66.42 on randomly sampled non-edge pairs to 58.77 on same-industry hard pairs. This suggests that industry membership and semantic similarity often serve as decision shortcuts.
Detailed comparisons with traditional and lightweight baselines, together with bootstrap uncertainty estimates, are provided in Appendix~\ref{app:additional-results}. The best zero-shot LLM outperforms all evaluated baselines, although the competitive supervised result indicates that task-specific decision patterns can also be learned effectively from
labeled examples.

\subsection{RQ2: Recovering Structured Collaboration Relations}

\textbf{Opportunity detection does not imply accurate recovery of the underlying relation structure. }Table~\ref{tab:structured-relation} reports complete per-model end-to-end results. Kimi-K2.7 obtains the highest Macro-F1 scores for collaboration type and role direction, reaching 60.23 and 67.71, despite achieving only 52.87 Macro-F1 for opportunity detection. In contrast, Qwen3.6-27B leads opportunity detection with 74.51 Macro-F1 but reaches only 51.05 and 56.11 for collaboration type and role direction.

\paragraph{Joint prediction remains substantially more difficult.} Kimi-K2.7 achieves the highest positive four-field exact match of 61.98\%, whereas Qwen3-2B obtains high overall exact match mainly by rejecting negative pairs and reaches only 4.65\% exact match on gold-positive instances. Detailed per-type results in Appendix~\ref{app:additional-results} show that Technology Transfer is particularly difficult and is frequently con fused with Supply and Production. Overall, current LLMs are considerably more reliable at identifying broad collaboration potential than at reconstructing its complete strength, mechanism, and role direction.

\paragraph{Where do structured-relation errors arise?}
Figure~\ref{fig:rq2-e2e-oracle} separates errors caused by missed
opportunities from errors in assigning relation attributes.  The two scores are close for most high-recall systems. Kimi-K2.7, for example, detects 97.42\% of gold-positive pairs and has a mean oracle--end-to-end gap of only
1.11 points across the three attributes.  Gemma4-26B, Qwen3.6-35B, GLM-5.2, and GLM-5.1 likewise have mean gaps below one point.  Their remaining errors therefore arise mainly after an opportunity has already been detected.  In
contrast, Qwen3-2B covers only 26.96\% of gold positives and its mean gap reaches 19.21 points, making missed detection the dominant bottleneck for that model.  Qwen3.6-27B also shows a non-negligible 5.52-point gap despite
leading opportunity detection overall. Complete joint and oracle-detection results are reported in
Appendix~E.2. Appendix~F further decomposes positive-pair errors
and analyzes systematic collaboration-type confusion and
role-direction bias.

\begin{figure}[ht!]
    \centering
    \includegraphics[width=\columnwidth]
    {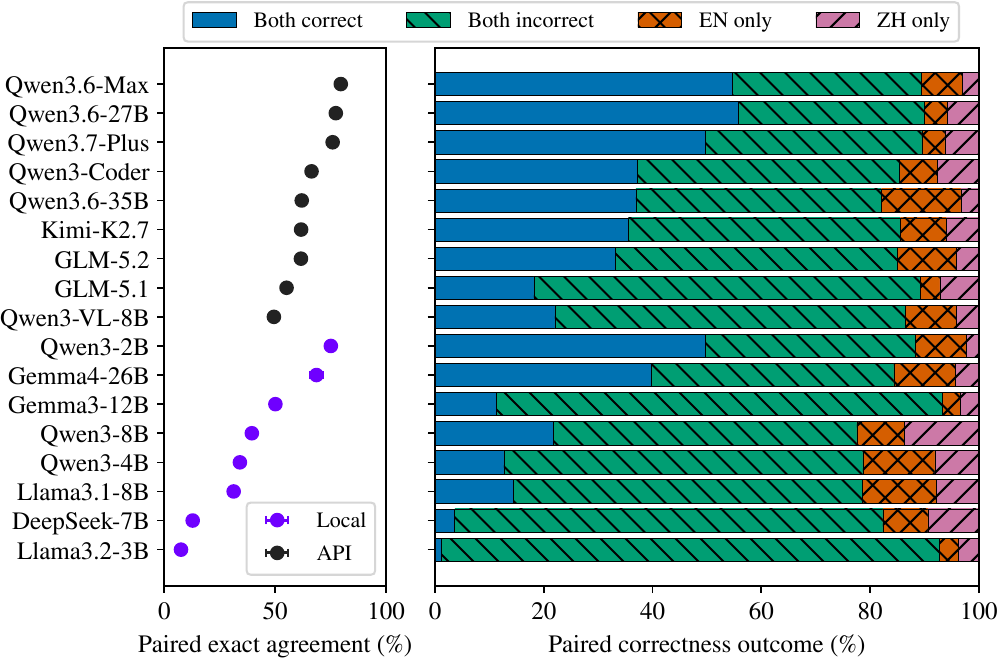}
    \caption{Paired Chinese--English prediction agreement and correctness outcomes. Paired agreement measures tuple identity and does not imply correctness.}
    \label{fig:rq3-paired}
\end{figure}

\begin{figure}[ht!]
    \centering
    \includegraphics[width=\columnwidth]
    {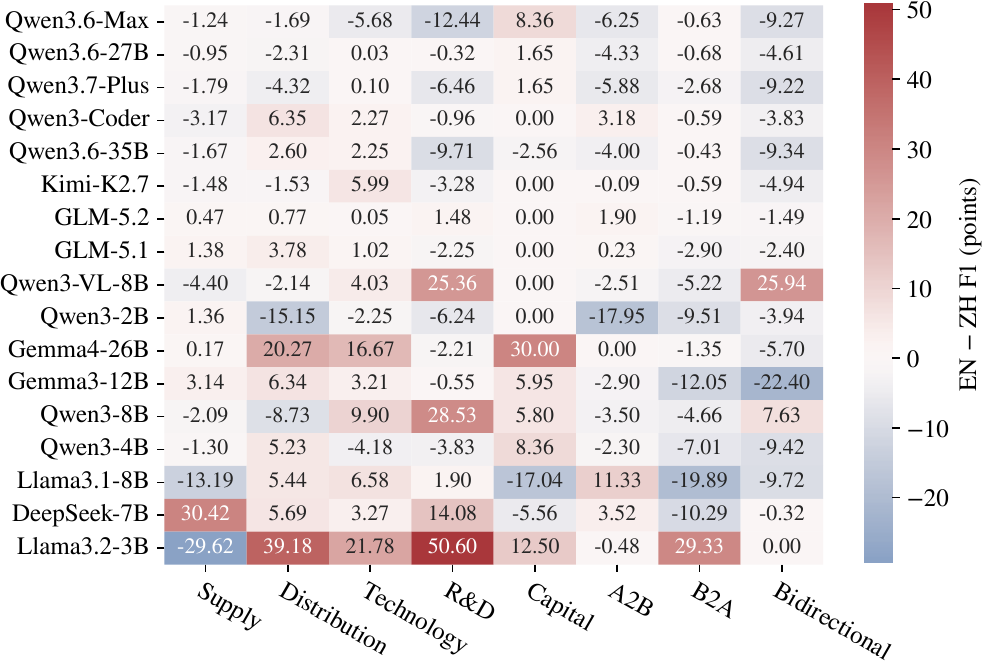}
    \caption{English-minus-Chinese class-level F1 differences for collaboration types and role directions. Positive values indicate higher performance under English input.}
    \label{fig:rq3-class-diff}
\end{figure}

\subsection{RQ3: Sensitivity to Input Language}

\paragraph{Input-language sensitivity.} Figure~\ref{fig:rq3-zh-en} shows that English input does not yield a consistent advantage. For example, four-field exact match increases by 11.34 points for Qwen3.6-35B but decreases by 4.96 points for Qwen3-8B, while Qwen3.6-27B remains relatively stable ($-1.43$ points). Complete model-level results are reported in Appendix Table~\ref{tab:crosslingual-full}. Figure~\ref{fig:rq3-paired} further shows that similar aggregate scores can conceal substantial instance-level changes. Moreover, high cross-language agreement does not necessarily indicate robust reasoning: Qwen3.6-Max produces identical predictions in 79.57\% of paired instances, but is correct in both languages on only 54.62\%, indicating that some errors are repeated across languages. 

\paragraph{Category-level variation.}
Figure~\ref{fig:rq3-class-diff} shows that language effects also vary across collaboration types and role directions. English improves some categories for particular models while degrading the same categories for others, and no relation category exhibits a consistent language advantage across all systems. Input-language sensitivity is therefore model-, instance-, and category-dependent rather than a uniform
performance shift.

\subsection{Reliability Analysis}

\paragraph{Valid outputs are not necessarily reliable.}
Most hosted models produce syntactically valid JSON, but their
cross-field consistency varies substantially. Qwen3-Coder achieves
the lowest violation rate of 0.21\%, whereas several models exceed 10\%. Self-reported confidence is also only partially informative: although Qwen3.6-27B, Qwen3.7-Plus, and Qwen3.6-Max exhibit large High--Low exact-match gaps, their high-confidence subsets still contain 31.86--39.64\% errors. Complete output-validity and
confidence-conditioned results are reported in Appendix~\ref{app:reliability}.

\section{Conclusion}

We introduce \textsc{FirmCORe}, a human-annotated benchmark for evaluating structured reasoning over inter-firm collaboration opportunities from weakly structured firm profiles. The task requires models to distinguish capability complementarity from superficial relatedness and jointly predict the existence, strength, type, and role direction of each opportunity. Experiments on 2,805 parallel Chinese--English samples show that current LLMs can identify broad collaboration potential but remain far less capable of reconstructing the full relational structure. They often overpredict opportunities based on mere relatedness, conflate distinct collaboration mechanisms, and misidentify the direction of resource flows. Language effects vary across models and instances, while high cross-lingual consistency may reflect the same errors being repeated across languages rather than genuinely robust reasoning. \textsc{FirmCORe} therefore provides a controlled testbed for diagnosing and improving profile-grounded, multidimensional inter-firm reasoning in future models.

\section{Limitations}

\textsc{FirmCORe} evaluates potential collaboration opportunities inferred from provided firm profiles, rather than realized partnerships or their commercial viability. Its pairwise setup does not address large-scale partner retrieval or ranking. Our experiments are further limited to zero-shot prompting and single deterministic runs, leaving fine-tuning, prompt sensitivity, tool use, and multi-run variability unexplored.


\bibliography{aaai2027}

@inproceedings{cao2024companykg,
  title={Companykg: A large-scale heterogeneous graph for company similarity quantification},
  author={Cao, Lele and von Ehrenheim, Vilhelm and Granroth-Wilding, Mark and Anselmo Stahl, Richard and McCornack, Andrew and Catovic, Armin and Cavalcanti Rocha, Dhiana Deva},
  booktitle={Proceedings of the 30th ACM SIGKDD Conference on Knowledge Discovery and Data Mining},
  pages={4816--4827},
  year={2024}
}

@article{yang2025emerging,
  title={Emerging industry classification based on BERT model},
  author={Yang, Baocheng and Zhang, Bing and Cutsforth, Kevin and Yu, Shanfu and Yu, Xiaowen},
  journal={Information Systems},
  volume={128},
  pages={102484},
  year={2025},
  publisher={Elsevier}
}

@article{kosasih2025towards,
  title={Towards trustworthy AI for link prediction in supply chain knowledge graph: a neurosymbolic reasoning approach},
  author={Kosasih, Edward Elson and Brintrup, Alexandra},
  journal={International Journal of Production Research},
  volume={63},
  number={6},
  pages={2268--2290},
  year={2025},
  publisher={Taylor \& Francis}
}

@article{li2025integrating,
  title={Integrating graph retrieval-augmented generation with large language models for supplier discovery},
  author={Li, Yunqing and Ko, Hyunwoong and Ameri, Farhad},
  journal={Journal of Computing and Information Science in Engineering},
  volume={25},
  number={2},
  pages={021010},
  year={2025},
  publisher={American Society of Mechanical Engineers}
}

@article{tu2024using,
  title={Using graph neural network to conduct supplier recommendation based on large-scale supply chain},
  author={Tu, Yuchun and Li, Wenxin and Song, Xiao and Gong, Kaiqi and Liu, Lu and Qin, Yunhao and Liu, Songsong and Liu, Ming},
  journal={International Journal of Production Research},
  volume={62},
  number={24},
  pages={8595--8608},
  year={2024},
  publisher={Taylor \& Francis}
}

@inproceedings{quan2024econlogicqa,
  title={Econlogicqa: A question-answering benchmark for evaluating large language models in economic sequential reasoning},
  author={Quan, Yinzhu and Liu, Zefang},
  booktitle={Findings of the Association for Computational Linguistics: EMNLP 2024},
  pages={2273--2282},
  year={2024}
}

@inproceedings{guan2026supchain,
  title={SupChain-Bench: Benchmarking Large Language Models for Real-World Supply Chain Management},
  author={Guan, Shengyue and Liu, Yihao and Cao, Lang},
  booktitle={Findings of the Association for Computational Linguistics: ACL 2026},
  pages={7526--7550},
  year={2026}
}

@article{hitt2000partner,
  title={Partner selection in emerging and developed market contexts: Resource-based and organizational learning perspectives},
  author={Hitt, Michael A and Dacin, M Tina and Levitas, Edward and Arregle, Jean-Luc and Borza, Anca},
  journal={Academy of Management journal},
  volume={43},
  number={3},
  pages={449--467},
  year={2000},
  publisher={Academy of Management Briarcliff Manor, NY 10510}
}

@article{furlotti2018fit,
  title={Fit for the task: Complementarity, asymmetry, and partner selection in alliances},
  author={Furlotti, Marco and Soda, Giuseppe},
  journal={Organization Science},
  volume={29},
  number={5},
  pages={837--854},
  year={2018},
  publisher={Informs}
}

@article{cabrera2021approach,
  title={An approach and decision support tool for forming Industry 4.0 supply chain collaborations},
  author={Cabrera, Sonia Cisneros and Pishchulov, Grigory and Sampaio, Pedro and Mehandjiev, Nikolay and Liu, Zixu and Kununka, Sophia},
  journal={Computers in Industry},
  volume={125},
  pages={1--16},
  year={2021},
  publisher={Elsevier BV}
}

@article{almahri2026enhancing,
  title={Enhancing supply chain visibility with knowledge graphs and large language models.},
  author={AlMahri, Sara and Xu, Liming and Brintrup, Alexandra},
  journal={International Journal of Production Research},
  volume={64},
  number={6},
  year={2026}
}

@article{zheng2025enhancing,
  title={Enhancing supply chain visibility with generative AI: an exploratory case study on relationship prediction in knowledge graphs},
  author={Zheng, Ge and Brintrup, Alexandra},
  journal={International Journal of Production Research},
  pages={1--23},
  year={2025},
  publisher={Taylor \& Francis}
}

@article{sun2025intercorprel,
  title={InterCorpRel-LLM: Enhancing Financial Relational Understanding with Graph-Language Models},
  author={Sun, Qianyou and Zheng, Jiexin and Jin, Bohan and Chen, Lihua and Peng, Yijie},
  journal={arXiv preprint arXiv:2510.09735},
  year={2025}
}

@article{mitsuhashi2009matching,
  title={A matching theory of alliance formation and organizational success: Complementarity and compatibility},
  author={Mitsuhashi, Hitoshi and Greve, Henrich R},
  journal={Academy of management journal},
  volume={52},
  number={5},
  pages={975--995},
  year={2009},
  publisher={American Society of Nephrology Briarcliff Manor, NY}
}

@article{mindruta2016two,
  title={A two-sided matching approach for partner selection and assessing complementarities in partners' attributes in inter-firm alliances},
  author={Mindruta, Denisa and Moeen, Mahka and Agarwal, Rajshree},
  journal={Strategic Management Journal},
  volume={37},
  number={1},
  pages={206--231},
  year={2016},
  publisher={Wiley Online Library}
}

@article{greve2013greener,
  title={Greener pastures: Outside options and strategic alliance withdrawal},
  author={Greve, Henrich R and Mitsuhashi, Hitoshi and Baum, Joel AC},
  journal={Organization Science},
  volume={24},
  number={1},
  pages={79--98},
  year={2013},
  publisher={INFORMS}
}

@article{davis2022machine,
  title={Machine learning-assisted industrial symbiosis: Testing the ability of word vectors to estimate similarity for material substitutions},
  author={Davis, Chris and Aid, Graham},
  journal={Journal of Industrial Ecology},
  volume={26},
  number={1},
  pages={27--43},
  year={2022},
  publisher={Wiley Online Library}
}

@inproceedings{krumdick2024bizbench,
  title={BizBench: A quantitative reasoning benchmark for business and finance},
  author={Krumdick, Michael and Koncel-Kedziorski, Rik and Lai, Viet Dac and Reddy, Varshini and Lovering, Charles and Tanner, Chris},
  booktitle={Proceedings of the 62nd Annual Meeting of the Association for Computational Linguistics (Volume 1: Long Papers)},
  pages={8309--8332},
  year={2024}
}

@inproceedings{matlin2025finance,
  title={Finance language model evaluation (flame)},
  author={Matlin, Glenn and Okamoto, Mika and Pardawala, Huzaifa and Yang, Yang and Chava, Sudheer},
  booktitle={Proceedings of the Fourth Workshop on Generation, Evaluation and Metrics (GEM$^2$)},
  pages={880--926},
  year={2025}
}

@article{bender2018data,
  title={Data statements for natural language processing: Toward mitigating system bias and enabling better science},
  author={Bender, Emily M and Friedman, Batya},
  journal={Transactions of the Association for Computational Linguistics},
  volume={6},
  pages={587--604},
  year={2018},
  publisher={MIT Press One Rogers Street, Cambridge, MA 02142-1209, USA journals-info~…}
}

@inproceedings{han2026mubench,
  title={Mubench: Assessment of multilingual capabilities of large language models across 61 languages},
  author={Han, Wenhan and Zhang, Yifan and Chen, Zhixun and Pechenizkiy, Mykola and Fang, Meng and Zheng, Yin and others},
  booktitle={Findings of the Association for Computational Linguistics: ACL 2026},
  pages={16163--16192},
  year={2026}
}


\clearpage

\appendix
\section{Candidate-Mining Details}
\label{app:candidate_mining}

For firm \(e_i\), let \(S_i\) and \(D_i\) denote its supply and
demand label sets, and let \(A_i=S_i\cup D_i\). We define
\begin{equation}
J(U,V)=
\begin{cases}
\dfrac{|U\cap V|}{|U\cup V|}, & |U\cup V|>0,\\
0, & \text{otherwise},
\end{cases}
\end{equation}
and compute
\begin{equation}
\operatorname{Sim}(i,j)=J(A_i,A_j),
\operatorname{Comp}(i,j)=\frac{J(D_i,S_j)+J(D_j,S_i)}{2}.
\end{equation}
These signals are used only for candidate mining and are not treated as gold labels.

\begin{table}[ht]
\centering
\small
\begin{tabular}{p{0.30\linewidth}p{0.62\linewidth}}
\hline
\textbf{Stratum} & \textbf{Criterion} \\
\hline
Strong-signal &
Upstream score \(\geq90\) \\

Borderline &
Upstream score in \([60,75)\) \\

Same-industry Hard &
Same industry,
\(\operatorname{Sim}(i,j)\geq0.03\), and
\(\operatorname{Comp}(i,j)\leq0.12\) \\

Random Non-edge &
No edge in the upstream candidate graph \\
\hline
\end{tabular}
\caption{Candidate-mining strata. Same-industry hard
pairs are not presumed to be negative.}
\label{tab:candidate_strata}
\end{table}

The upstream candidate graph and scores are generated by
DeepSeek-V4-Flash from firm profiles and published
collaboration-opportunity information. We retain at most
\(3{,}000\) pairs per stratum, remove cross-stratum and reversed
duplicates, and randomly assign pairs satisfying multiple
criteria to one stratum, as shown in Table~\ref{tab:candidate_strata}. For scored edges, the Firm A/Firm B order follows the fixed randomized order.

Qwen3.7-Max is subsequently used only to screen for
potential label leakage, privacy risks, and data-quality issues.
At most \(750\) pairs per stratum are retained for human
annotation. Candidate strata, upstream scores, and source-record
identifiers are hidden from annotators.

\section{Chinese–English Parallel Translation}
\label{app:zh-en}

The original firm profiles and annotations in \textsc{FirmCORe} were collected and constructed in Chinese. To support cross-lingual evaluation, we translated the full dataset into English using a controlled translation and quality-assurance protocol. The objective was not to produce a literal word-by-word translation, but to preserve the business meaning, entity identity, relational direction, and evidential boundaries of each original record. The resulting English version contains the same 2,805 annotated firm pairs as the Chinese version and preserves all gold labels without modification.

\paragraph{Profile-level translation.}
Each firm profile was decomposed into five fields: entity name, industry, core business, products or services, and firm summary. Translation was performed at the profile level rather than independently at each occurrence. After removing repeated profiles, 2,256 unique profiles were identified and assigned a single English representation. The same translation was then reused whenever the profile appeared in multiple firm pairs. This procedure prevents a Chinese entity or business description from receiving inconsistent translations across different examples.

Business descriptions were translated into natural professional English while preserving the scope of the source text. In particular, we avoided introducing capabilities, products, qualifications, or business relationships that were not explicitly supported by the Chinese profile. Long Chinese nominal expressions were reorganized when necessary to improve readability, but their semantic content was not expanded. Enumeration boundaries were also retained so that separate products, services, and technical capabilities were not incorrectly merged.

\paragraph{Entity-name normalization and verification.}
Entity names required special treatment because the dataset contains listed companies, private firms, public institutions, universities, government departments, branches, stores, research centers, projects, brands, and other non-standard organizational labels. For internationally recognized firms and institutions, we preferentially used the English name published on the entity's official website or in official corporate materials. Examples include \emph{Alibaba Group}, \emph{Tencent}, \emph{Huawei}, \emph{Chinese Academy of Sciences}, and \emph{Tsinghua University}.

When no verifiable official English name was available, we produced a standardized English rendering based on the Chinese name and its organizational form. Such renderings were not treated as official legal names. For local firms whose names contained distinctive Chinese brand terms, transliteration was retained where a literal translation would obscure the entity identity. Descriptive translation was used for generic institutional names, such as research centers, administrative offices, branches, and service platforms.

We recorded the provenance and confidence of entity-name translations using three review levels. \emph{Verified} names were supported by official English-language sources. \emph{Review recommended} was assigned when the translation was linguistically reliable but the official English name could not be independently confirmed, or when the record referred to a branch, department, project, or non-corporate organization. \emph{Review required} was assigned when the Chinese source name was abbreviated, malformed, generic, combined multiple entities, or otherwise insufficient for unique entity resolution.

\paragraph{Domain terminology.}
A controlled terminology glossary was maintained for recurring concepts in corporate registration, manufacturing, supply chains, technology transfer, and public-sector administration. Terms were translated according to their functional meaning rather than by surface lexical correspondence. 

The five collaboration labels were also translated using fixed expressions throughout the dataset: \emph{Supply and Production Collaboration}, \emph{Marketing and Distribution Collaboration}, \emph{R\&D and Co-development Collaboration}, \emph{Licensing and Technology Transfer Collaboration}, and \emph{Capital and Equity Collaboration}. Direction labels were preserved explicitly. Thus, \textsc{A2B} indicates that Firm A provides products, services, technology, capabilities, or other resources to Firm B, whereas \textsc{B2A} indicates the reverse direction. No direction label was inferred or altered during translation.

\paragraph{Preservation of source uncertainty.}
The Chinese profiles occasionally contained incomplete names, conflicting industry descriptions, historical company names, or inconsistencies between the listed industry and business summary. Translation was not used to silently repair these source-level issues. Instead, the English version preserves the available information and, when necessary, explicitly signals that the source profile is inconsistent or that the legal entity cannot be uniquely identified. This design prevents translation from artificially improving the informational quality of one language version and ensures that Chinese--English comparisons remain valid.

\paragraph{Quality assurance.}
The translated data underwent several automatic and manual checks. We verified that every original pair had a corresponding English record, that repeated Chinese profiles received identical English translations, and that all collaboration labels, scores, and role directions were unchanged. We additionally checked for missing English fields, residual Chinese characters in translated fields, inconsistent English names for the same Chinese entity, and malformed profile structures. High-risk entity names and source inconsistencies were retained in dedicated review fields. These procedures resulted in a traceable bilingual dataset in which each English profile can be directly aligned with its Chinese source, translation status, and review note.

\section{Data Statement}
\label{app:data-statement}

\paragraph{Data source and acquisition date.}
The source data underlying \textsc{FirmCORe} was obtained from the Xunfu platform through an operational data export generated in June 2026. The export contains core organization and business records, supplementary profile information, upstream candidate matching relations, and AI generated labels. We use these records to construct unified organization profiles, intermediate supply and demand tables, candidate pairs, source blind annotation sets, and the final gold standard benchmark. Due to commercial access restrictions and licensing constraints, we release only the final gold labeled benchmark data.

\paragraph{Data sources by field.}
The raw platform fields used in \textsc{FirmCORe} include organization names, selected industry and organization type attributes, and available business or contact related metadata. Derived fields include profile length, label overlap, business similarity, supply and demand complementarity, and shared label sets. These fields are computed using deterministic scripts. Other intermediate fields, including text summaries, AI generated labels, supply and demand labels, candidate relation scores, and explanatory fields, originate from upstream AI or platform systems. \textsc{FirmCORe} is therefore not an unmodified collection of corporate registration records.

\paragraph{Entity matching, deduplication, and filtering.}
Organization profiles are constructed primarily by aligning names across platform tables, supplemented by available organization identifiers. Duplicate field values are merged, records with missing or invalid organization names are removed, and duplicate profiles are deduplicated. Profiles that contain insufficient business information for meaningful pairwise assessment are also excluded. During candidate pair construction, we remove self pairs, duplicate unordered pairs, reversed duplicates, pairs containing invalid profiles, and pairs containing fields that could reveal candidate generation signals. The current entity matching process relies mainly on organization names and deterministic rules rather than a comprehensive entity resolution system.

\paragraph{Candidate pair construction and representativeness.}
\textsc{FirmCORe} is not uniformly sampled from all possible organization pairs. Candidate pairs are deliberately mined from four strata: strong signal upstream pairs, borderline signal upstream pairs, same industry hard pairs, and random non-edge pairs. We first construct a larger candidate pool for each stratum, then apply AI based quality screening and source blind human annotation. Up to 750 pairs are selected from each stratum, producing an annotation pool of 3,000 pairs. After complete tuple aggregation and disagreement resolution, 2,805 instances receive definitive gold labels. This stratified design covers decision regions ranging from obvious to difficult cases and prevents the benchmark from being dominated by trivial random negatives.

\paragraph{Privacy and sensitive information.}
The original platform exports contain fields that are unsuitable for direct public release, including phone numbers, email addresses, authentication tokens, session identifiers, national identification numbers, instant messaging accounts, and other potentially sensitive metadata. \textsc{FirmCORe} does not release the original exports. Human annotation and model evaluation are based on minimized organization level business profiles that exclude direct contact details, authentication information, and internal platform fields. We also remove fields such as \texttt{reason}, \texttt{action\_plan}, and \texttt{raw\_json}, since they may contain sensitive content or directly reveal upstream system judgments. These measures reduce unnecessary exposure but do not replace independent legal, privacy, and data governance review. Any public release should therefore be limited to the minimum information required to reproduce the benchmark task.

\paragraph{Label semantics and intended use.}
A positive label in \textsc{FirmCORe} indicates that, under the annotation guidelines, the two provided organization profiles contain sufficient evidence to support a plausible collaboration opportunity. Positive instances are further labeled with opportunity strength, primary collaboration type, and role direction. These labels do not indicate that the organizations have collaborated in practice. A negative label indicates that the available profiles do not provide sufficient evidence for a collaboration opportunity under the annotation guidelines. Since relevant capabilities or constraints may be absent from the profiles, a negative label should not be interpreted as evidence that collaboration is impossible in practice. \textsc{FirmCORe} is intended to evaluate profile based relation reasoning and structured relation prediction.

\paragraph{Known biases and limitations.}
\textsc{FirmCORe} inherits several potential biases. First, the source data primarily reflects Chinese commercial and institutional contexts. The English evaluation set is translated from the same Chinese profiles rather than independently collected from naturally occurring English organization profiles. Second, the gold labels contain fewer positive than negative instances. Positive instances are also concentrated in strong opportunities, the Supply and Production category, and one displayed role direction. These distributions may affect aggregate accuracy and class specific performance. We therefore report Macro-F1 and per class metrics in addition to accuracy. Finally, organization profiles reflect business information available at a particular point in time. They may become outdated as products, capabilities, and organizational roles change.

\section{Prompts}
\label{app:prompt}

We use language-aligned prompts for the Chinese and English evaluation sets. The two prompts share the same task definition, label space, consistency constraints, and JSON output schema.
The Chinese prompt requires a brief rationale in Chinese, while the English prompt requires a brief rationale in English. We provide the complete English prompt below.

\begin{promptbox}

You are an annotator specializing in inter-firm collaboration
opportunities. Given the profiles of two industry entities,
Company A and Company B, determine whether the provided
information supports a potential collaboration opportunity
between them.

Base your judgment strictly on the provided profiles. Do not
introduce external knowledge or assume facts that are not stated
or reasonably supported by the text.

Output JSON only. Do not output Markdown, additional
explanations, or any content outside the JSON object.

The annotation fields are defined as follows.

\textbf{1. has\_opportunity}

\begin{itemize}
    \item \texttt{Yes}: The profiles provide sufficient evidence
    for a plausible collaboration opportunity between A and B.
    \item \texttt{No}: The profiles do not provide sufficient
    evidence for a clear collaboration or capability-complementarity
    relationship.
\end{itemize}

\textbf{2. opportunity\_score}

\begin{itemize}
    \item \texttt{1}: Weak opportunity. Cooperation is plausible,
    but the relationship is indirect, the evidence is limited, or
    the collaboration chain is relatively long.
    \item \texttt{2}: Strong opportunity. One party's products,
    services, technologies, channels, capabilities, resources, or
    capital can relatively directly support the other party.
\end{itemize}

\textbf{3. cooperation\_type}

Select exactly one of the following:

\begin{itemize}
    \item \texttt{Supply and Production Collaboration}
    \item \texttt{Marketing and Distribution Collaboration}
    \item \texttt{Licensing and Technology Transfer Collaboration}
    \item \texttt{R\&D and Co-development Collaboration}
    \item \texttt{Capital and Equity Collaboration}
\end{itemize}

\textbf{4. role\_direction}

\begin{itemize}
    \item \texttt{A2B}: Under the selected cooperation type,
    Company A primarily provides Company B with the relevant
    products, services, technologies, capabilities, resources,
    channels, or capital.
    \item \texttt{B2A}: Under the selected cooperation type,
    Company B primarily provides Company A with the relevant
    products, services, technologies, capabilities, resources,
    channels, or capital.
    \item \texttt{Bidirectional}: The opportunity primarily involves
    mutual exchange, joint development, or two-way value creation.
\end{itemize}

\textbf{5. confidence}

\begin{itemize}
    \item \texttt{1}: Low confidence.
    \item \texttt{2}: High confidence.
\end{itemize}

\textbf{6. reason}

Provide one concise English sentence explaining the evidence
for the judgment.

\textbf{Consistency requirements}

\begin{itemize}
    \item If \texttt{has\_opportunity = No}, then
    \texttt{opportunity\_score}, \texttt{cooperation\_type}, and
    \texttt{role\_direction} must all be \texttt{None}.
    \item If \texttt{has\_opportunity = Yes}, then
    \texttt{opportunity\_score} must be \texttt{1} or \texttt{2},
    and neither \texttt{cooperation\_type} nor
    \texttt{role\_direction} may be \texttt{None}.
    \item A strong opportunity must be supported by a relatively
    direct product, service, technology, channel, capability,
    resource, or capital relationship.
    \item Business similarity, shared industry membership, or
    lexical overlap alone is insufficient evidence for a positive
    label.
\end{itemize}

\textbf{Company A}

\texttt{\{object\_a\_profile\}}

\textbf{Company B}

\texttt{\{object\_b\_profile\}}

Return exactly one JSON object using the following schema:

\medskip
\noindent
{\ttfamily\footnotesize\raggedright
\{\\
\hspace*{1em}"has\_opportunity": "Yes / No",\\
\hspace*{1em}"opportunity\_score": "1 / 2",\\
\hspace*{1em}"cooperation\_type":
\hspace*{2em}"Supply and Production Collaboration / \hspace*{2em}Marketing and Distribution Collaboration / \hspace*{2em}Licensing and Technology Transfer Collaboration / \hspace*{2em}R\&D and Co-development Collaboration / \hspace*{2em}Capital and Equity Collaboration",
\hspace*{1em}"role\_direction":\\
"A2B / B2A / Bidirectional",\\
\hspace*{1em}"confidence": "1 / 2",\\
\hspace*{1em}"reason": "One concise English sentence"\\
\}
}

\end{promptbox}

\section{Additional Results}
\label{app:additional-results}

\subsection{Baselines and Detection Uncertainty}
\label{app:baselines}

\begin{table*}[t]
\centering
\small
\caption{Comparison of opportunity-detection baselines on the Chinese benchmark.}
\label{tab:opportunity-baselines}
\begin{tabular}{lccccc}
\toprule
Baseline & Macro-F1 & Positive P & Positive R & Positive F1 & MCC \\
\midrule
Always-No & 39.57 & 0.00 & 0.00 & 0.00 & 0.00 \\
Majority-Prior & 39.57 & 0.00 & 0.00 & 0.00 & 0.00 \\
Always-Yes & 25.66 & 34.51 & 100.00 & 51.31 & 0.00 \\
\midrule
LSA-Cosine & 43.67 & 33.55 & 64.46 & 44.13 & -2.84 \\
\midrule
Com.-Threshold & 46.43 & 67.29 & 7.44 & 13.40 & 13.73 \\
Industry-Rule & 33.29 & 9.11 & 7.33 & 8.13 & -33.12 \\
\midrule
Upstream-Score & 67.29 & 64.78 & 46.18 & 53.92 & 36.37 \\
\midrule
TF-IDF-Cosine & 47.71 & 37.17 & 74.38 & 49.57 & 8.36 \\
J.S. & 47.18 & 34.00 & 54.75 & 41.95 & -1.21 \\
\midrule
L.S.C & \underline{71.69} & 59.89 & 70.04 & 64.57 & 43.92 \\
\midrule
Best zero-shot LLM (Qwen3.6-27B) & \textbf{74.51} & 59.99 & 84.71 & 70.24 & 52.25 \\
\bottomrule
\end{tabular}
\end{table*}

Table~\ref{tab:opportunity-baselines} compares the best evaluated zero-shot LLM, Qwen3.6-27B, with trivial priors, lexical and embedding-based similarity methods, rule-based signals, the upstream scoring system, and a
lightweight supervised classifier. The trivial baselines expose the effect of class imbalance: Always-Yes reaches 100\% positive recall but only 25.66 Macro-F1, whereas Always-No and the majority-tuple baseline obtain 39.57 Macro-F1 without identifying any positive instance. Unsupervised similarity
and rule-based methods remain below 48 Macro-F1. Industry similarity performs particularly poorly, confirming that shared industry membership alone is not a reliable proxy for collaboration potential. Supply--demand complementarity attains relatively high precision (67.29\%) but very low recall (7.44\%), indicating that direct label overlap is overly conservative
and misses implicit capability complementarity.

The upstream score provides a stronger reference at 67.29 Macro-F1, and the lightweight supervised classifier reaches 71.69. Qwen3.6-27B achieves the best overall result, with 74.51 Macro-F1, 52.25 MCC, and a comparatively balanced positive precision and recall of 59.99\% and 84.71\%. These results
show that opportunity detection benefits from semantic reasoning beyond lexical similarity and manually defined matching signals, while the competitive supervised result indicates that task-specific decision patterns can also be learned effectively from labeled examples.

\begin{figure}[t]
  \centering
  \includegraphics[width=\columnwidth]{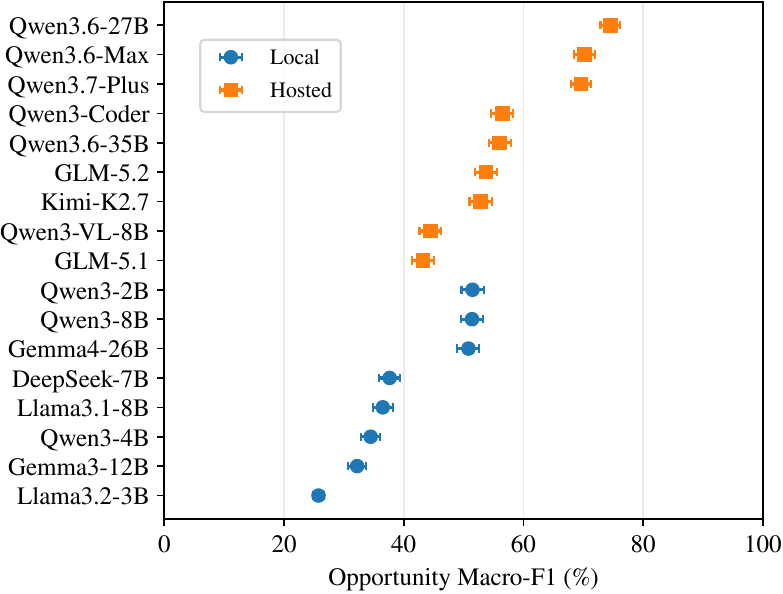}
  \caption{Opportunity-detection Macro-F1 with 95\% instance-bootstrap
  confidence intervals on the Chinese benchmark.}
  \label{fig:app-rq1-ci}
\end{figure}

Figure~\ref{fig:app-rq1-ci} complements the point estimates with
instance-bootstrap uncertainty. Qwen3.6-27B remains the strongest
opportunity detector at 74.51 Macro-F1 (95\% CI: 72.89--76.07), followed by Qwen3.6-Max and Qwen3.7-Plus. These intervals quantify instance-level uncertainty, but they do not capture prompt sensitivity or run-to-run variation because each model was evaluated once under deterministic decoding.

\subsection{Joint and Oracle Structured Prediction}
\label{app:structured-results}

\begin{table*}[htbp]
\centering
\small
\begin{tabular}{lrrrrrr}
\toprule
Model & Overall & Pos. EM & 95\% CI & Cov. & Cond. Attr. EM & MCF \\
\midrule
Qwen3-8B & 35.47 & 47.62 & [44.42, 50.72] & 93.80 & 50.77 & 3.00 \\
Gemma4-26B & 34.47 & 47.11 & [44.01, 50.31] & 97.93 & 48.10 & 3.09 \\
Llama3.1-8B & 22.25 & 44.11 & [41.01, 47.21] & 98.97 & 44.57 & 3.04 \\
Qwen3-4B & 20.71 & 43.49 & [40.39, 46.59] & 97.93 & 44.41 & 3.13 \\
Gemma3-12B & 14.69 & 30.89 & [28.10, 33.78] & 99.90 & 30.92 & 2.83 \\
Llama3.2-3B & 4.92 & 14.15 & [11.98, 16.43] & 99.69 & 14.20 & 2.34 \\
Deepseek-7B & 12.98 & 13.12 & [11.05, 15.29] & 94.73 & 13.85 & 2.45 \\
Qwen3-2B & 51.98 & 4.65 & [3.41, 5.99] & 26.96 & 17.24 & 0.70 \\
\midrule
Kimi-K2.7 & 41.50 & 61.98 & [58.78, 64.98] & 97.42 & 63.63 & 3.38 \\
Qwen3-Coder & 44.92 & 56.10 & [53.00, 59.30] & 89.98 & 62.34 & 3.10 \\
Qwen3.6-Max & 57.61 & 52.27 & [49.17, 55.37] & 89.46 & 58.43 & 3.01 \\
Qwen3.7-Plus & 55.90 & 51.65 & [48.45, 54.86] & 91.84 & 56.24 & 3.05 \\
Qwen3.6-35B & 40.21 & 50.31 & [47.11, 53.51] & 97.62 & 51.53 & 3.13 \\
GLM-5.2 & 37.22 & 47.11 & [43.90, 50.31] & 97.21 & 48.46 & 3.08 \\
Qwen3.6-27B & 61.57 & 45.14 & [42.05, 48.24] & 84.71 & 53.29 & 2.73 \\
GLM-5.1 & 25.35 & 38.95 & [35.95, 41.94] & 98.86 & 39.39 & 2.94 \\
Qwen3-VL-8B & 26.17 & 37.81 & [34.81, 40.91] & 96.80 & 39.06 & 2.85 \\
\bottomrule
\end{tabular}
\caption{Joint structured prediction performance. Positive four-field exact
match requires opportunity existence, strength, collaboration type, and role
direction to be simultaneously correct on every gold-positive instance.
Detection-conditioned three-attribute exact match is computed only for
gold-positive instances predicted as opportunities and is reported together
with positive coverage.}
\label{tab:positive-joint-exact-match}
\begin{minipage}{0.98\textwidth}\footnotesize
\textit{Note.}
All values except MCF are percentages.
Overall denotes four-field exact match over all instances.
Pos.\ EM denotes four-field exact match on gold-positive instances.
Cov.\ denotes positive coverage, i.e., the proportion of gold-positive instances
predicted as opportunities.
Cond.\ Attr.\ EM denotes detection-conditioned three-attribute exact match.
MCF denotes mean correct fields.
\end{minipage}
\end{table*}

\paragraph{Joint prediction.}
Table~\ref{tab:positive-joint-exact-match} evaluates whether opportunity
existence, strength, collaboration type, and role direction are recovered simultaneously. Kimi-K2.7 achieves the highest positive four-field exact match of 61.98\%, followed by Qwen3-Coder at 56.10\%. In contrast, Qwen3-2B obtains an overall exact match of 51.98\% but only 4.65\% exact match on gold-positive instances because it detects only 26.96\% of them. Overall exact match can therefore be dominated by correct rejection of negative pairs and should be interpreted together with positive exact match and positive coverage.

\begin{table*}[t]
\centering
\small
\caption{Oracle-detection conditional performance.}
\label{tab:oracle-conditional}
\begin{tabular}{lccccc}
\toprule
Model & Coverage & Strength Macro-F1 & Type Macro-F1 & Direction Macro-F1 & Oracle--E2E Gap \\
\midrule
Qwen3-2B & 26.96 & 46.40 & 29.37 & 27.85 & 19.21 \\
Qwen3-8B & 93.80 & 47.99 & 49.89 & 54.87 & 2.27 \\
Gemma4-26B & 97.93 & 46.50 & 54.46 & 67.23 & 0.78 \\
DeepSeek-7B & 94.73 & 47.72 & 32.32 & 37.24 & 0.94 \\
Llama3.1-8B & 98.97 & 46.17 & 43.70 & 52.51 & 0.48 \\
Qwen3-4B & 97.93 & 51.27 & 57.70 & 55.54 & 0.58 \\
Gemma3-12B & 99.90 & 45.30 & 44.45 & 61.04 & 0.02 \\
Llama3.2-3B & 99.69 & 38.92 & 14.56 & 22.29 & 0.04 \\
\midrule
Qwen3.6-27B & 84.71 & 48.68 & 57.23 & 62.43 & 5.52 \\
Qwen3.6-Max & 89.46 & 52.80 & 57.17 & 63.43 & 4.32 \\
Qwen3.7-Plus & 91.84 & 48.82 & 60.94 & 70.30 & 2.73 \\
Qwen3-Coder & 89.98 & 49.15 & 49.86 & 62.94 & 3.02 \\
Qwen3.6-35B & 97.62 & 48.62 & 55.63 & 64.35 & 0.88 \\
GLM-5.2 & 97.21 & 50.71 & 55.61 & 60.76 & 0.99 \\
Kimi-K2.7 & 97.42 & 53.81 & 61.71 & 68.85 & 1.11 \\
Qwen3-VL-8B & 96.80 & 42.56 & 41.15 & 55.56 & 0.93 \\
GLM-5.1 & 98.86 & 42.59 & 57.15 & 64.11 & 0.51 \\
\bottomrule
\end{tabular}
\begin{minipage}{0.98\textwidth}\footnotesize Oracle includes only gold-positive instances correctly detected as Yes. Coverage is opportunity recall on gold-positive instances; the gap is the mean oracle minus E2E Macro-F1 across three attributes.\end{minipage}
\end{table*}

\paragraph{Detection-conditioned analysis.}
Table~\ref{tab:oracle-conditional} reports relation-attribute performance conditional on correctly detecting a gold-positive pair. Coverage must be read together with the conditional scores: a high oracle score at low
coverage characterizes only the subset already detected and is not a
directly comparable end-to-end ranking. For most high-recall systems, the oracle--end-to-end gap is small, indicating that their remaining errors arise mainly when assigning strength, collaboration type, or role direction after
an opportunity has already been detected. Missed detection is instead the dominant bottleneck for Qwen3-2B. Detailed type-confusion and direction-bias analyses are provided in Appendix~\ref{app:error-analysis}.

\subsection{Complete Cross-Lingual Results}
\label{app:crosslingual-results}

\begin{table*}[t]
\centering
\small
\caption{Paired Chinese--English evaluation. Paired Agreement measures tuple
identity rather than correctness.}
\label{tab:crosslingual-full}
\begin{tabular}{lccccccc}
\toprule
Model & ZH EM & EN EM & $\Delta$EM & Paired Agreement & Both Correct & ZH Only & EN Only \\
\midrule
Qwen3-2B & 51.98 & 59.22 & +7.24 & 75.08 & 49.73 & 2.25 & 9.48 \\
Gemma4-26B & 44.15 & 51.11 & +6.95 & 68.60 & 39.83 & 4.32 & 11.28 \\
Gemma3-12B & 14.69 & 14.47 & -0.21 & 50.09 & 11.27 & 3.42 & 3.21 \\
Qwen3-8B & 35.47 & 30.52 & -4.96 & 39.47 & 21.78 & 13.69 & 8.73 \\
Qwen3-4B & 20.71 & 25.92 & +5.20 & 34.08 & 12.66 & 8.06 & 13.26 \\
Llama3.1-8B & 22.25 & 27.95 & +5.70 & 31.27 & 14.40 & 7.84 & 13.55 \\
DeepSeek-7B & 12.98 & 11.94 & -1.03 & 12.80 & 3.64 & 9.34 & 8.31 \\
Llama3.2-3B & 4.92 & 4.67 & -0.25 & 7.52 & 1.18 & 3.74 & 3.49 \\
\midrule
Qwen3.6-Max & 57.61 & 62.21 & +4.60 & 79.57 & 54.62 & 2.99 & 7.59 \\
Qwen3.6-27B & 61.57 & 60.14 & -1.43 & 77.33 & 55.83 & 5.74 & 4.31 \\
Qwen3.7-Plus & 55.90 & 53.87 & -2.03 & 75.90 & 49.66 & 6.24 & 4.21 \\
Qwen3-Coder & 44.92 & 44.21 & -0.71 & 66.38 & 37.29 & 7.63 & 6.92 \\
Qwen3.6-35B & 40.21 & 51.55 & +11.34 & 61.96 & 36.93 & 3.28 & 14.62 \\
Kimi-K2.7 & 41.50 & 44.10 & +2.60 & 61.68 & 35.58 & 5.92 & 8.52 \\
GLM-5.2 & 37.22 & 44.10 & +6.88 & 61.60 & 33.16 & 4.06 & 10.94 \\
GLM-5.1 & 25.35 & 21.82 & -3.53 & 55.12 & 18.18 & 7.17 & 3.64 \\
Qwen3-VL-8B & 26.17 & 31.34 & +5.17 & 49.41 & 22.03 & 4.14 & 9.30 \\
\bottomrule
\end{tabular}

\begin{minipage}{0.98\textwidth}
\footnotesize
Scores are percentages; $\Delta\mathrm{EM}=\mathrm{EN}-\mathrm{ZH}$.
Paired Agreement measures predicted-tuple identity across the Chinese and
English versions and does not imply correctness.
Complete pairs have $N=2{,}805$, although usable pair counts may vary by model.
Correctness denotes four-field exact match.
\end{minipage}
\end{table*}

Table~\ref{tab:crosslingual-full} provides the complete model-level
statistics underlying the cross-lingual analysis. English input produces neither a universal gain nor a universal decline. Models with similar Chinese and English aggregate exact-match scores may nevertheless show substantial asymmetry between the ZH-only and EN-only subsets. Moreover, paired agreement is consistently higher than the proportion of instances answered correctly in both languages, demonstrating that part of the apparent cross-language stability is attributable to the same incorrect tuple being repeated across the two input conditions.

\subsection{Robustness to A/B Firm-Order Permutation}
\label{app:ab-permutation}

To test whether predictions depend on presentation order, we construct a direction-balanced variant of the Chinese test set. The variant retains all 2,805 firm pairs and preserves opportunity existence, opportunity strength, and collaboration type. We reverse the displayed order only for gold-positive pairs with a unidirectional relation. The original test set contains 157 A2B and 700 B2A instances. Using a fixed random seed, we swap 427 pairs, including 78 original A2B instances and 349 original B2A
instances, yielding an approximately balanced distribution of 428 A2B and 429 B2A pairs. Under ideal equivariance, a swapped pair should preserve opportunity existence, strength, and type while reversing only A2B and B2A. Predictions for unswapped pairs should remain unchanged.

\begin{table*}[t]
\centering
\caption{Prediction stability under A/B firm-order permutation. All rates
are percentages. The balanced set contains the same 2,805 firm pairs as the
original Chinese test set, with 427 gold-positive unidirectional pairs shown
in reversed order.}
\label{tab:ab-permutation}

\small
\setlength{\tabcolsep}{5.0pt}
\renewcommand{\arraystretch}{1.08}

\textbf{(a) Prediction equivariance}

\begin{tabular}{@{}lrrrrr@{}}
\toprule
Model
& Exact Eq. $\uparrow$
& Direction Eq. $\uparrow$
& Overturned $\downarrow$
& Severe $\downarrow$
& Swapped Pos. Overturned $\downarrow$ \\
\midrule
Qwen3.6-Max  & \textbf{91.37} & \textbf{94.44} & \textbf{8.63}  & \textbf{6.74}  & 26.00 \\
Qwen3-VL-8B  & 88.63 & 93.05 & 11.37 & 8.09  & 39.81 \\
Qwen3.6-35B  & 86.35 & 89.77 & 13.65 & 8.48  & 34.19 \\
Qwen3.7-Plus & 81.21 & 91.27 & 18.79 & 16.93 & 28.10 \\
Qwen3.6-27B  & 81.89 & 89.55 & 18.11 & 16.33 & 36.07 \\
GLM-5.1      & 51.98 & 73.23 & 48.02 & 38.86 & 41.92 \\
Qwen3-Coder  & 80.25 & 84.99 & 19.75 & 13.80 & \textbf{22.25} \\
Kimi-K2.7    & 62.82 & 77.79 & 37.18 & 31.12 & 25.53 \\
GLM-5.2      & 63.74 & 79.04 & 36.26 & 29.84 & 38.17 \\
\midrule
\textbf{Average}
& \textbf{76.47}
& \textbf{85.90}
& \textbf{23.53}
& \textbf{18.91}
& \textbf{32.45} \\
\bottomrule
\end{tabular}

\textbf{(b) Changes in four-field exact-match correctness}

\begin{tabular}{@{}lrrrrr@{}}
\toprule
Model
& Correct$\rightarrow$Wrong $\downarrow$
& Wrong$\rightarrow$Correct $\uparrow$
& Net Correction $\uparrow$
& Both Correct $\uparrow$
& Both Wrong $\downarrow$ \\
\midrule
Qwen3.6-Max  & 1.96 & 2.03 & 0.07 & 55.69 & 40.32 \\
Qwen3-VL-8B  & \textbf{1.11} & 2.35 & 1.25 & 25.06 & 71.48 \\
Qwen3.6-35B  & 1.78 & 2.82 & 1.03 & 38.43 & 56.97 \\
Qwen3.7-Plus & 2.71 & 3.03 & 0.32 & 53.19 & 41.07 \\
Qwen3.6-27B  & 1.64 & 4.67 & 3.03 & \textbf{59.93} & \textbf{33.76} \\
GLM-5.1      & 3.10 & 5.38 & 2.28 & 22.25 & 69.27 \\
Qwen3-Coder  & 2.00 & 6.38 & 4.39 & 42.92 & 48.70 \\
Kimi-K2.7    & 5.49 & 7.38 & 1.89 & 36.01 & 51.12 \\
GLM-5.2      & 2.35 & \textbf{7.42} & \textbf{5.06} & 34.87 & 55.37 \\
\midrule
\textbf{Average}
& \textbf{2.46}
& \textbf{4.61}
& \textbf{2.15}
& \textbf{40.93}
& \textbf{52.01} \\
\bottomrule
\end{tabular}

\begin{minipage}{0.98\textwidth}
\footnotesize
\textit{Note.}
Exact equivariance requires the complete structured prediction to remain
unchanged for unswapped pairs and to undergo only the expected A2B/B2A
reversal for swapped pairs. Direction equivariance evaluates only the
direction field. \textit{Severe} denotes changes involving opportunity
existence, opportunity strength, or collaboration type rather than direction
alone. Correctness-transition rates and the \textit{Both Correct} and
\textit{Both Wrong} columns are computed over all 2,805 pairs.
\textit{Swapped Pos. Overturned} is computed only over the 427 reversed
gold-positive unidirectional pairs. Net Correction equals
Wrong$\rightarrow$Correct minus Correct$\rightarrow$Wrong.
\end{minipage}
\end{table*}

Table~\ref{tab:ab-permutation} reports results for the nine models with
predictions on both the original and balanced variants. Averaged across
models, exact equivariance is 76.47\% and direction equivariance is 85.90\%.
Thus, 23.53\% of predictions fail to preserve the expected structured
equivalence, and 18.91\% change not only direction but also opportunity
existence, strength, or collaboration type. Order sensitivity is therefore
not confined to the directional label.

\begin{figure}[t]
\centering
\includegraphics[width=\columnwidth]
{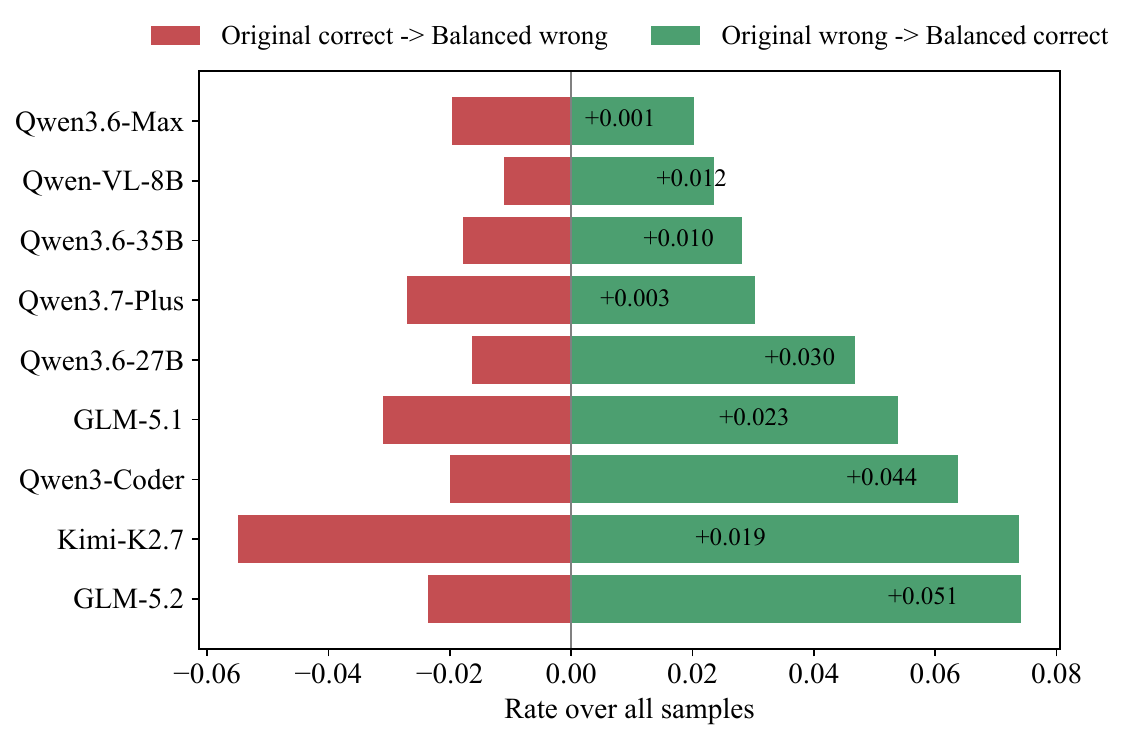}
\caption{Changes in four-field exact-match correctness after A/B firm-order
permutation. Leftward and rightward bars denote Correct$\rightarrow$Wrong and
Wrong$\rightarrow$Correct transitions, respectively; annotations report net
correction rates.}
\label{fig:ab-permutation-flips}
\end{figure}

A/B permutation does not systematically overturn previously correct
predictions. As shown in Figure~\ref{fig:ab-permutation-flips}, an average of
2.46\% of instances change from correct to incorrect, whereas 4.61\% change
from incorrect to correct, for a net improvement of 2.15 points. The balanced
variant should therefore be interpreted as a stress test for order
sensitivity rather than as evidence against the main conclusions drawn from
the original benchmark.

\subsection{Output Validity and Confidence Reliability}
\label{app:reliability}

\begin{table*}[ht!]
\centering
\caption{Output validity and self-reported confidence reliability on the Chinese benchmark.}
\label{tab:output-reliability}
\small
\setlength{\tabcolsep}{3.2pt}
\begin{tabular}{l*{10}{S[table-format=+3.2]}}
\toprule
 & \multicolumn{5}{c}{Output validity} & \multicolumn{5}{c}{Confidence reliability} \\
\cmidrule(lr){2-6}\cmidrule(lr){7-11}
Model & {Strict JSON $\uparrow$} & {Parse $\uparrow$} & {Missing $\downarrow$} & {Invalid $\downarrow$} & {Violation $\downarrow$} & {High Cov.} & {EM@High} & {EM@Low} & {Gap} & {High Err. $\downarrow$} \\
\midrule
Qwen3-2B & 100.00 & 100.00 & 0.46 & 0.00 & 0.46 & 42.14 & 36.55 & 63.22 & -26.67 & 63.45 \\
Qwen3-8B & 0.00 & 96.54 & 24.21 & 3.46 & 20.75 & 69.73 & 33.13 & 46.14 & -13.01 & 66.87 \\
Gemma4-26B & 99.43 & 100.00 & 2.64 & 0.00 & 2.64 & 90.34 & 38.12 & 0.37 & 37.75 & 61.88 \\
DeepSeek-7B & 0.00 & 99.07 & 10.20 & 0.93 & 9.27 & 77.86 & 5.82 & 31.03 & -25.22 & 94.18 \\
Llama3.1-8B & 100.00 & 100.00 & 5.13 & 0.00 & 5.13 & 92.62 & 16.44 & 95.12 & -78.69 & 83.56 \\
Qwen3-4B & 0.00 & 99.04 & 16.93 & 0.96 & 15.97 & 95.58 & 21.67 & 0.00 & 21.67 & 78.33 \\
Gemma3-12B & 0.00 & 100.00 & 8.81 & 0.00 & 8.81 & 90.45 & 11.79 & 42.16 & -30.38 & 88.21 \\
Llama3.2-3B & 99.32 & 99.32 & 12.73 & 0.68 & 12.05 & 85.42 & 5.72 & 0.26 & 5.46 & 94.28 \\
\midrule
Qwen3.6-27B & 100.00 & 100.00 & 1.43 & 0.00 & 1.43 & 90.20 & 68.14 & 1.09 & 67.05 & 31.86 \\
Qwen3.6-Max & 99.89 & 99.89 & 1.99 & 0.11 & 1.89 & 86.24 & 65.40 & 8.83 & 56.57 & 34.60 \\
Qwen3.7-Plus & 100.00 & 100.00 & 1.03 & 0.00 & 1.03 & 92.55 & 60.36 & 0.48 & 59.88 & 39.64 \\
Qwen3-Coder & 100.00 & 100.00 & 0.21 & 0.00 & 0.21 & 98.86 & 44.32 & 96.88 & -52.55 & 55.68 \\
Qwen3.6-35B & 99.96 & 99.96 & 13.12 & 0.04 & 13.08 & 97.54 & 40.94 & 11.76 & 29.17 & 59.06 \\
GLM-5.2 & 100.00 & 100.00 & 8.13 & 0.00 & 8.13 & 60.78 & 56.72 & 7.00 & 49.72 & 43.28 \\
Kimi-K2.7 & 99.96 & 99.96 & 2.71 & 0.04 & 2.67 & 83.71 & 49.23 & 1.75 & 47.48 & 50.77 \\
Qwen3-VL-8B & 98.97 & 98.97 & 16.19 & 1.03 & 15.15 & 98.93 & 26.41 & 100.00 & -73.59 & 73.59 \\
GLM-5.1 & 100.00 & 100.00 & 15.29 & 0.00 & 15.29 & 28.27 & 51.45 & 15.06 & 36.39 & 48.55 \\
\bottomrule
\end{tabular}
\begin{minipage}{0.98\textwidth}\footnotesize Correctness is four-field exact match. Gap is EM@High minus EM@Low. Strict JSON is measured on untouched raw output; Parse is measured after deterministic normalization. Missing, Invalid, and Violation are error rates over all predictions. ``--" indicates zero coverage or unavailable conditioning. Confidence is a two-level self-report, not a calibrated probability.\end{minipage}
\end{table*}

Table~\ref{tab:output-reliability} jointly reports raw JSON validity,
normalized parsing success, missing-field and invalid-label rates,
cross-field constraint violations, and confidence-conditioned exact match.
Most hosted models produce syntactically valid outputs, but comparable
parsing success does not imply comparable structural consistency.
Qwen3-Coder records the lowest violation rate at 0.21\%, followed by
Qwen3-2B (0.46\%), Qwen3.7-Plus (1.03\%), Qwen3.6-27B (1.43\%), and
Qwen3.6-Max (1.89\%). Common errors include retaining non-empty relation
attributes after predicting \textsc{No} and omitting required fields after
predicting \textsc{Yes}.

Self-reported confidence provides useful risk stratification for some models
but is not a calibrated probability. Qwen3.6-27B, Qwen3.7-Plus, and
Qwen3.6-Max show High--Low exact-match gaps of 67.05, 59.88, and 56.57
points, respectively, yet their High-confidence subsets still contain
31.86--39.64\% errors. Interpretation also depends strongly on coverage:
some models assign nearly all instances to the High-confidence subset,
making the Low-confidence estimate unstable. Confidence should therefore be
treated as a model-specific diagnostic and reported together with
High-confidence coverage.


\section{Error Analyses}
\label{app:error-analysis}

\paragraph{Evaluation alignment.}
All analyses in this section use the instance-level prediction table aligned to the 2,805 gold pair identifiers. Duplicate outputs, API failures, missing predictions, invalid labels, and cross-field violations are retained and handled explicitly. Cross-lingual statistics are computed only on firm pairs for which both Chinese and English predictions are available for the same model.

\begin{figure}[t]
    \centering
    \includegraphics[width=\columnwidth]{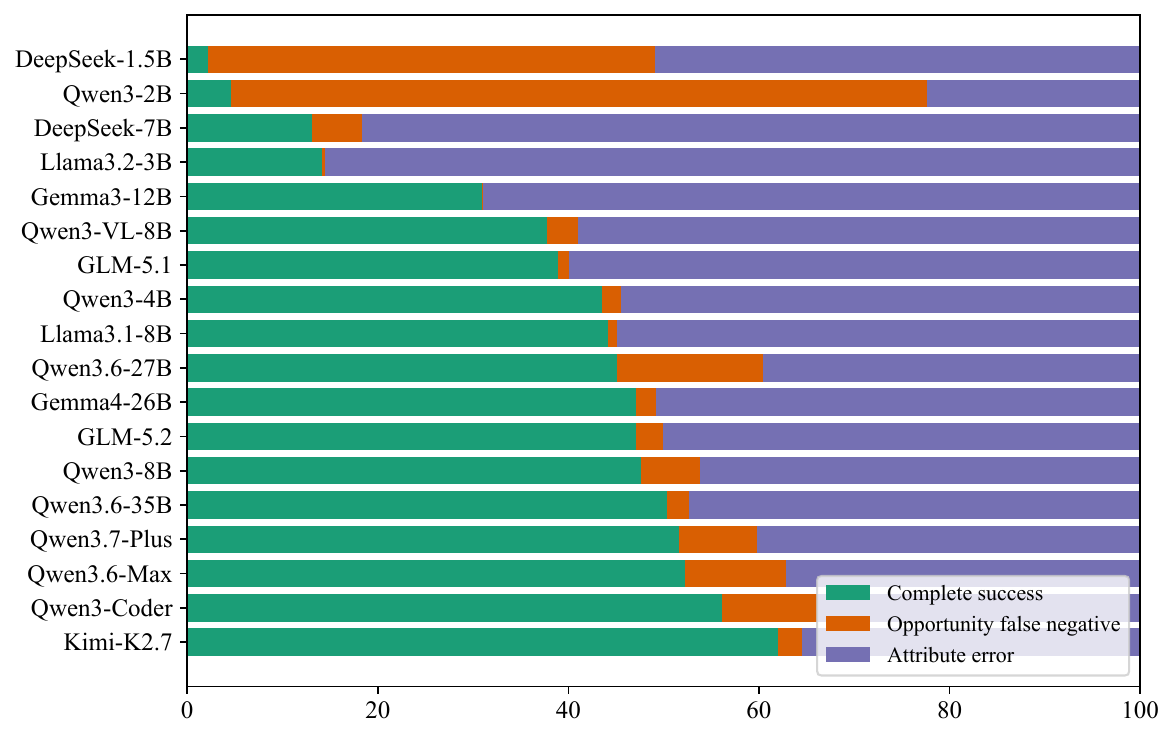}
    \caption{Prediction outcomes on the 968 gold-positive Chinese instances. Each prediction belongs to one of three mutually exclusive categories: (i) complete tuple correct; (ii) opportunity false negative, where the gold label is Yes but the model predicts No; or (iii) attribute error after a positive opportunity prediction, where the model predicts Yes but at least one of strength, collaboration type, or role direction is incorrect, missing, or invalid.}
    \label{fig:app-error-decomposition}
\end{figure}

\subsection{Error Decomposition}
\label{app:error-decomposition}

Figure~\ref{fig:app-error-decomposition} decomposes predictions on the 968 gold-positive pairs into complete tuple correctness, opportunity false negatives, and attribute errors conditional on a positive opportunity prediction. Kimi-K2.7 obtains the highest positive four-field exact match, recovering the complete tuple for 600 of 968 pairs (62.0\%). It predicts the opportunity-existence label as No for 25 gold-positive pairs (2.6\%). For another 343 pairs (35.4\%), it correctly predicts Yes but returns at least one incorrect, missing, or invalid relation attribute. Qwen3.6-27B is the strongest opportunity detector in the main evaluation, yet its positive four-field exact match is 437/968 (45.1\%): 148 gold-positive pairs (15.3\%) are predicted as No, while 383 (39.6\%) receive a positive opportunity prediction with at least one erroneous relation attribute. By contrast, Qwen3-2B predicts No for 707 gold-positive pairs (73.0\%) and recovers the complete tuple for only 45 (4.6\%). Opportunity false negatives therefore dominate the errors of low-recall models, whereas high-recall systems are limited primarily by collaboration-type and role-direction errors after predicting that an opportunity exists.

\subsection{Systematic Failure Modes}
\label{app:hard-failure-modes}

\begin{table*}[t]
\centering
\small
\setlength{\tabcolsep}{4pt}
\caption{Aggregate failure modes on the Chinese benchmark. Counts are model--instance decisions aggregated over 17 models. Denominators include all eligible decisions for the corresponding task.}
\label{tab:app-failure-summary}
\begin{tabular}{p{3.4cm}p{5.0cm}p{3.0cm}p{5.2cm}}
\toprule
Failure mode & Operational criterion & Frequency & Main pattern \\
\midrule
Unsupported opportunity prediction & Gold opportunity is No, but the model predicts Yes. & 23,621/34,903 (67.7\%) & Broad industrial or lexical relatedness is frequently treated as sufficient evidence for a concrete collaboration interface. \\
Opportunity false negative & Gold opportunity is Yes, but the model predicts No. & 1,903/18,392 (10.3\%) & Cross-field evidence for complementarity is sometimes insufficiently integrated, leading the model to assign the negative opportunity label. \\
Valid but incorrect type & The gold label is Yes, the model predicts Yes, and the returned valid type differs from the gold type. & 4,495/18,392 (24.4\%) & Predictions collapse toward frequent or semantically adjacent mechanisms, especially Supply and Production. \\
Valid but incorrect direction & The gold label is Yes, the model predicts Yes, and the returned valid direction differs from the gold direction. & 5,092/18,392 (27.7\%) & Errors reflect both majority-class attraction and model-specific overprediction of A2B or Bidirectional relations. \\
\bottomrule
\end{tabular}
\end{table*}

\paragraph{Relatedness mistaken for complementarity.}
Unsupported opportunity predictions are the most frequent error at the model--instance level. Among the 1,836 gold-negative pairs misclassified by at least one Chinese model, 1,773 (96.6\%) are misclassified by three or more models. Moreover, 18,551 of the 23,621 false-positive decisions (78.5\%) are assigned High confidence. At the pair level, both same-industry hard pairs (644/644) and random non-edges (626/627) attract at least one false positive. The recurrence across models indicates that broad industry proximity, shared terminology, or market adjacency often triggers a plausible collaboration narrative even when the supplied profiles do not identify a sufficiently specific product, technology, channel, capability, or capital flow.

\paragraph{Opportunity false negatives on implicit complementarity.}
Across the 18,392 gold-positive model--instance decisions, models predict No in 1,903 cases (10.3\%), producing opportunity false negatives. At the pair level, 804 of 968 positive pairs receive at least one false-negative prediction, and 209 receive false-negative predictions from at least three models. These errors are not confined to low-signal candidates: at least one model predicts No for 299/358 borderline pairs (83.5\%) and 347/447 strong-signal pairs (77.6\%). The pattern suggests that models do not consistently integrate evidence distributed across products, services, trading capabilities, channels, and business roles. When complementarity is expressed indirectly or with limited lexical overlap, the model may assign the negative opportunity label even though the supplied profiles support a positive relation.

\begin{figure}[t]
    \centering
    \includegraphics[width=\columnwidth]{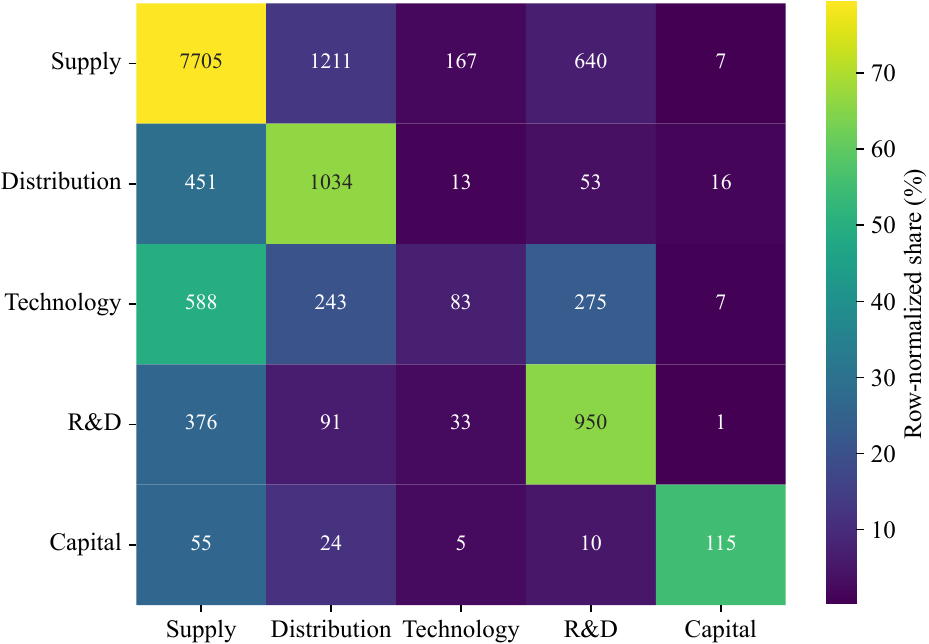}
    \caption{Collaboration-type confusion aggregated over 17 models on gold-positive Chinese instances for which the model predicts \textsc{Yes} and returns a valid collaboration type. Cell annotations report aggregated prediction counts, while cell colors encode row-normalized percentages. Gold-positive instances predicted as \textsc{No}, together with predictions containing invalid or missing type outputs, are excluded from the matrix.}
    \label{fig:app-type-confusion}
\end{figure}

\paragraph{Collaboration-type confusion.}
Type-related errors have three distinct forms. Across all gold-positive model--instance decisions, models predict No in 1,903 cases. In another 1,458 cases, they predict Yes but return an invalid or missing type, and in 4,495 cases, they predict Yes and return a valid type that differs from the gold label. Figure~\ref{fig:app-type-confusion} isolates the third category. Technology Transfer is particularly unstable: among valid type predictions for Technology instances, only 6.5\% are correct, while 48.9\% are assigned Supply and Production and 23.4\% R\&D and Co-development. Distribution and R\&D are also frequently mapped to Supply and Production (27.6\% and 25.6\%, respectively). These errors suggest that models often recognize a broad basis for collaboration but fail to distinguish the resource being exchanged: a delivered product or service, an existing technology, or a jointly developed capability. 

\begin{figure}[t]
    \centering
    \includegraphics[width=\columnwidth]{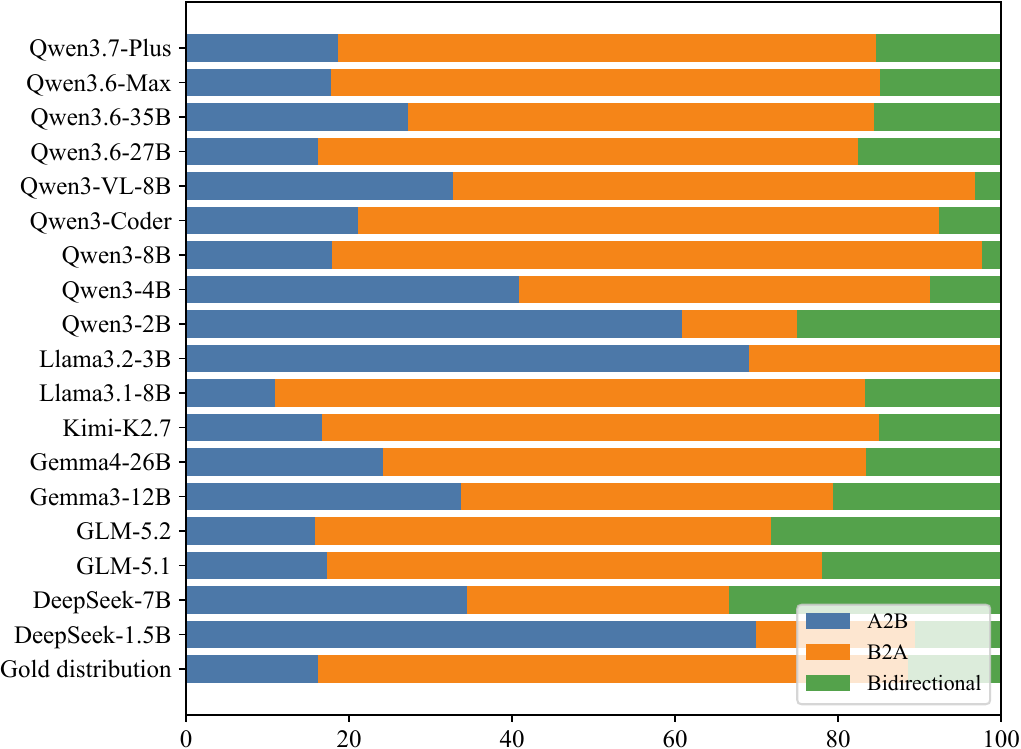}
    \caption{Distribution of valid role-direction predictions on gold-positive Chinese instances for which the model predicts Yes. Models are ordered by their predicted B2A share, and the gold distribution is included as a reference. Gold-positive instances predicted as No and positive predictions with invalid or missing directions are excluded from the normalization.}
    \label{fig:app-direction-distribution}
\end{figure}

\paragraph{Role-direction bias and reversal.}
The gold direction distribution is highly imbalanced: 157 pairs are A2B, 700 B2A, and 111 Bidirectional. An always-B2A classifier therefore reaches 72.31\% accuracy without performing pair-specific direction reasoning. Figure~\ref{fig:app-direction-distribution} shows that model biases are heterogeneous. Qwen3-8B assigns 79.9\% of its valid direction predictions to B2A, compared with 72.3\% in the gold data. In contrast, DeepSeek-7B predicts Bidirectional for 33.4\% of valid outputs, nearly three times the gold share of 11.5\%, while Llama3.2-3B assigns 69.1\% to A2B and never predicts Bidirectional. Kimi-K2.7 attains the strongest direction Macro-F1 (67.71) and 77.38\% accuracy, only 5.07 points above the majority-direction baseline. 

\subsection{Cross-Lingual and Confidence-Related Failures}
\label{app:crosslingual-confidence-failures}

\begin{table*}[t]
\centering
\small
\setlength{\tabcolsep}{5pt}
\caption{Representative paired Chinese--English outcomes. Values are percentages of usable paired instances. ``Identical wrong'' denotes the same incorrect four-field tuple in both languages.}
\label{tab:app-crosslingual-errors}
\begin{tabular}{lrrrrrr}
\toprule
Model & $N$ & Both correct & Identical wrong & Different wrong & ZH only & EN only \\
\midrule
Qwen3.6-27B & 2,805 & 55.8 & 21.5 & 12.6 & 5.7 & 4.3 \\
Qwen3.6-Max & 2,805 & 54.6 & 25.0 & 9.8 & 3.0 & 7.6 \\
Qwen3.7-Plus & 2,805 & 49.7 & 26.2 & 13.7 & 6.2 & 4.2 \\
Kimi-K2.7 & 2,805 & 35.6 & 26.1 & 23.9 & 5.9 & 8.5 \\
Gemma3-12B & 2,805 & 11.3 & 38.8 & 43.3 & 3.4 & 3.2 \\
Llama3.2-3B & 2,805 & 1.2 & 6.3 & 85.2 & 3.7 & 3.5 \\
\bottomrule
\end{tabular}
\begin{minipage}{0.98\textwidth}\footnotesize
\end{minipage}
\end{table*}

High cross-lingual agreement can reflect shared errors rather than robust reasoning. As shown in Table~\ref{tab:app-crosslingual-errors}, the same incorrect tuple is produced in both languages for 21.5\% of Qwen3.6-27B pairs, 25.0\% of Qwen3.6-Max pairs, and 26.2\% of Qwen3.7-Plus pairs. For Gemma3-12B, identical wrong tuples account for 38.8\% of pairs, while a further 43.3\% receive different incorrect tuples. These outcomes distinguish two failure modes: language-invariant reasoning shortcuts and language-sensitive decision changes. Neither pattern, by itself, identifies translation quality as the cause.

Self-reported confidence is similarly informative only as a model-specific diagnostic. Qwen3.6-27B labels 90.20\% of Chinese predictions as High confidence, yet 31.86\% of that subset is incorrect under four-field exact match. The corresponding high-confidence error rates are 34.60\% for Qwen3.6-Max and 39.64\% for Qwen3.7-Plus. Thus, the binary confidence label separates risk for some models but remains far from a calibrated probability of correctness.

\subsection{Representative Failure Cases}
\label{app:representative-failure-cases}

\begin{table*}[t]
\centering
\footnotesize
\setlength{\tabcolsep}{2.5pt}
\caption{Representative diagnostic cases. Evidence summaries are derived only from the supplied firm profiles.}
\label{tab:app-failure-cases}
\begin{tabular}{p{1.7cm}p{2.4cm}p{4.9cm}p{3.0cm}p{3.7cm}}
\toprule
Pair & Pattern & Key profile evidence & Gold / representative prediction & Diagnostic interpretation \\
\midrule
HSAI\_00165 & Unsupported opportunity & \textbf{A:} telecommunications services, broadband, IoT, cloud, and data centers. \textbf{B:} ICT infrastructure, network equipment, enterprise solutions, cloud, and smart devices. & No / Yes--Strong--Supply--B2A & Broad ICT overlap is treated as a specific supply interface, although neither profile states a concrete demand relationship. \\
MLAI\_01557 & Opportunity false negative & \textbf{A:} retail of furniture, home-improvement materials, appliances, bathroom products, and lighting. \textbf{B:} broad import--export trade in agricultural products, machinery, textiles, and other goods. & Yes--Strong--Distribution--B2A / No & The model must connect a broad trading capability to a downstream retail channel despite limited direct product overlap. \\
HSAI\_00848 & Type and direction confusion & \textbf{A:} media distribution, advertising, brand promotion, and digital marketing. \textbf{B:} digital-marketing platforms, ad-delivery systems, analytics, and related technical services. & Yes--Strong--Technology--B2A / Yes--Strong--Distribution--Bidirectional & Commercial-channel cues obscure both the technical mechanism and the annotated resource-flow direction. \\
RNEG\_01776 & High-confidence type error & \textbf{A:} property sales, leasing, and consulting. \textbf{B:} real-estate development, investment, asset management, and property management. & Yes--Strong--Supply--A2B / Yes--Strong--Distribution--A2B & The direction is preserved, but service provision is conflated with channel distribution under High confidence. \\
\bottomrule
\end{tabular}
\end{table*}

The cases in Table~\ref{tab:app-failure-cases} illustrate three recurring boundaries. First, profile-level relatedness can support a plausible narrative without supplying the specific interface required by the gold label. Second, implicit opportunities may require combining a broad capability with a downstream business role, making them difficult to recover from lexical overlap alone. Third, semantically adjacent mechanisms remain difficult to separate even when the model correctly predicts the opportunity-existence label as Yes. In HSAI\_00848, the representative prediction changes both the primary collaboration mechanism and the resource-flow direction. RNEG\_01776 further shows that High confidence does not prevent a narrower type error: the direction is correct, but Supply and Production is replaced by Marketing and Distribution. These cases show that identifying broad collaboration opportunities is easier than reconstructing the complete relational structure.

\end{document}